\pdfoutput=1

\documentclass[11pt]{article}

\usepackage[]{acl}

\usepackage{times}
\usepackage{latexsym}
\usepackage{multirow}
\usepackage{booktabs}
\usepackage{graphicx}
\usepackage{amssymb}
\usepackage{makecell}
\usepackage{amsmath, bm}
\usepackage{longtable}
\usepackage{subfigure}
\usepackage{caption}
\usepackage{enumerate}
\usepackage{enumitem}

\usepackage{algorithm}
\usepackage{algorithmic}
\usepackage{longtable}
\usepackage{xspace}
\usepackage{comment}

\usepackage{cuted}
\usepackage{url}

\usepackage{amssymb}

\usepackage{tcolorbox}
\tcbuselibrary{listingsutf8}
\newtcolorbox{callout}[1][]{colback=yellow!10!white,colframe=yellow!50!black,fonttitle=\bfseries,title=#1}
\newtcolorbox{graycallout}[1][]{colback=gray!10,colframe=gray!70,fonttitle=\bfseries,title=#1}
\newtcolorbox{infobox}[1][]{colback=gray!10, colframe=gray!70, fonttitle=\bfseries, title=#1}

\definecolor{green1}{RGB}{215,228,190}
\definecolor{red1}{RGB}{242,217,217}
\definecolor{orange1}{RGB}{255,241,221}
\definecolor{blue1}{RGB}{222,240,250}

\usepackage{amsmath}
\DeclareMathOperator*{\argmax}{arg\,max}
\DeclareMathOperator*{\argmin}{arg\,min}

\makeatletter

\newcommand{\Rmnum}[1]{\expandafter\@slowromancap\romannumeral #1@}
\makeatother

\usepackage[T1]{fontenc}

\usepackage[utf8]{inputenc}

\usepackage{microtype}

\usepackage{inconsolata}

\title{Aligning VLM Assistants with Personalized Situated Cognition}

\author{
\textbf{Yongqi Li$^{1}$}, 
\textbf{Shen Zhou$^{1}$}, 
\textbf{Xiaohu Li$^{1}$}, 
\textbf{Xin Miao$^{1}$}, 
\textbf{Jintao Wen$^{1}$}, 
\textbf{Mayi Xu$^{1}$}, 
\textbf{Jianhao Chen$^{1,2}$}, \\
\textbf{Birong Pan$^{1}$}, 
\textbf{Hankun Kang$^{1}$}, 
\textbf{Yuanyuan Zhu$^{1,*}$}, 
\textbf{Ming Zhong$^{1}$}, 
\textbf{Tieyun Qian$^{1,2,}$\thanks{{ }{ }Corresponding authors.}}\\
        $^1$ School of Computer Science, Wuhan University, China \\ 
        $^2$ Zhongguancun Academy, Beijing, China\\
        \texttt{\{liyongqi,yyzhu,qty\}@whu.edu.cn}}

\begin{document}
\maketitle
\begin{abstract}

Vision-language models (VLMs) aligned with general human objectives, such as being harmless and hallucination-free, have become valuable assistants of humans in managing visual tasks. However, people with diversified backgrounds have different cognition even in the same situation. Consequently, they may have personalized expectations for VLM assistants. This highlights the urgent need to \textit{align VLM assistants with personalized situated cognition} for real-world assistance.
To study this problem, we first simplify it by characterizing individuals based on the sociological concept of \textit{Role-Set}.
Then, we propose to evaluate the individuals' actions to examine whether the personalized alignment is achieved.
Further, we construct a benchmark named \textbf{PCogAlignBench}, which includes 18k instances and 20 individuals with different Role-Sets. 
Finally, we present a framework called \textbf{PCogAlign}, which constructs a cognition-aware and action-based reward model for personalized alignment.
Experimental results and human evaluations demonstrate the reliability of the PCogAlignBench and the effectiveness of our proposed PCogAlign. We will open-source the constructed benchmark and code at \url{https://github.com/NLPGM/PCogAlign}.

\end{abstract}

\section{Introduction}
Recently, significant progress has been made in aligning vision-language models (VLMs)~\cite{hurst-2024-gpt4o} with humans through visual instruction tuning~\cite{liu-2024-llava}, visual preference optimization~\cite{sun-2023-RLHFV,zhao-2023-beyond}, and safety alignment~\cite{zong-2024-safetyVSFT,wang-2024-crossmodalitysafetyalignment}. These studies focus on general alignment goals, e.g., following human instructions, reducing visual hallucinations, and aligning with human values. 
Based on the success of these techniques, VLMs have become essential assistants for humans in managing visual tasks~\cite{yin-2023-MLLMsurvey}.

\begin{figure}[t!]
\centering
\includegraphics[width=0.4\textwidth]{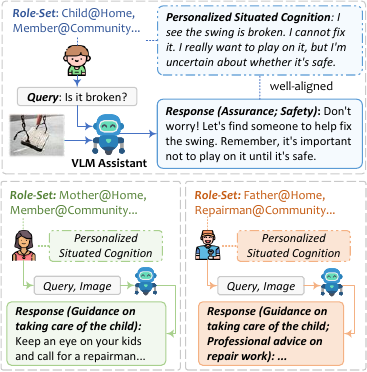}
\caption{Individuals with different Role-Sets exhibit personalized situated cognition when encountering the same ``\textit{broken swing}'' visual scene, resulting in personalized expectations for the VLM assistants' responses.}
\label{fig:introduction_figure}
\end{figure}

However, even in the same situation, people with diversified backgrounds might have personalized situated cognition~\footnote{Referencing the Situated Cognition Theory~\cite{brown-1989-SituatedCognition}, we define the \textit{situated cognition} discussed in this paper as: 1) cognition of the visual scene state; 2) cognition of body and mind states; 3) cognition of the appropriate next action.}. Consequently, they may have personalized expectations for the VLM assistants' responses.
To illustrate this, we present a ``\textit{broken swing}'' visual scene as an example in Figure~\ref{fig:introduction_figure}. 
When faced with the ``broken swing'' scenario, a child may feel unsure and worried due to their limited ability to address such issues. In this context, the VLM assistant is expected to offer reassurance and basic safety guidance, ensuring that the child understands not to play on the swing for their safety and easing their concerns. In contrast, a repairman possesses the expertise to fix the problem. Therefore, the VLM assistant should provide professional advice on repairing the swing, enabling the repairman to take appropriate actions.

The personalized situated cognition and personalized expectation reveal that current VLM assistants, which provide a one-size-fits-all response, are inadequate.
Therefore, advancing research to align VLM assistants with personalized situated cognition is crucial. 
This will enable them to be more effective collaborators, capable of meeting personalized expectations from diverse individuals.

To study this problem, we first need to consider how to define diverse individuals. Given that human diversity stems from numerous factors, such as age and socioeconomic status, which are almost impossible to operationalize in experiments, we have to make necessary simplifications.
Based on the concept of the \textit{Role-Set} from Role Theory~\cite{goffman-2023-presentation}, we characterize each individual as a set of ``\textit{Role@Location}'' components, such as ``Child@Home, Member@Community'' as illustrated in Figure~\ref{fig:introduction_figure}. Individuals with different Role-Sets exhibit personalized situated cognitions and therefore expect personalized responses.

Based on the above definition, we construct a benchmark named PCogAlignBench to investigate how to align VLM assistants with personalized situated cognition. 
Specifically, we select 8 social locations~\cite{oldenburg-1989-great} where individuals might engage and define 3 to 5 roles for each location (32 roles in total).
Based on certain combination constraints, we define 20 Role-Sets, each of which consists of 5 ``\textit{Role@Location}'' combinations.
Then, we automatically collect 12k training samples and 6k test samples.
The test samples undergo rigorous quality control through human annotators to ensure evaluation reliability.
Each sample includes a Role-Set, an image, and a query posed by the individual.
Additionally, every test sample includes an ``\textit{oracle guidance}'', which describes ``the characteristics of the expected personalized response''.


Furthermore, we propose a framework named PCogAlign for this problem, which consists of three steps.
1) Estimating the individual's situated cognition and optimal action. 
2) Sampling multiple personalized responses via cooperative agents.
3) Constructing and utilizing a cognition-aware and action-based reward model to iteratively select the optimal response.
The selected optimal responses are expected to align well with individuals' personalized situated cognition and assist individuals in making optimal actions, and will be used as feedback for alignment training.
Experimental results in various settings consistently prove the superiority of PCogAlign compared to baselines.


In all, our study makes three main contributions.

\textbf{1) Novel Task.} We introduce a novel task that aims to align VLM assistants with personalized situated cognition to meet diverse expectations.

\textbf{2) New Benchmark.} We present a new benchmark named PCogAlignBench to explore this novel task, which consists of 18k samples.

\textbf{3) Novel Methodology.} We propose a novel framework named PCogAlign as a baseline for this task. Comprehensive experimental results and analysis validate its effectiveness.



\section{Related Work}

\paragraph{Alignment of Vision-Language Models}
Existing techniques for aligning vision-language models (VLMs) with humans can be categorized into three types based on their alignment objectives: 1) \textit{visual instruction tuning}~\cite{liu-2024-llava,liu-2024-improvedLlava} to enable VLMs to follow human instructions in visual tasks; 2) \textit{visual preference optimization} to mitigate visual hallucinations in VLMs' responses~\cite{sun-2023-RLHFV,zhu-2024-selfVisualAlign,yang-2025-VDPOKey,wang-2024-mdpo,xing-2024-efuf,yu-2024-rlhfv,yu-2024-rlaifv}; and 3) \textit{safety fine-tuning} to align VLMs with human values and prevent harmful outputs~\cite{chen-2024-dress,shi-2024-AssMLLMValues,zhang-2024-SPAVL}.
The success of these alignment techniques has enabled VLMs to be effective assistants for humans in tasks involving visual scenes~\cite{yin-2024-mllmSurvey}. 
However, considering the diversity of individuals' cognition, even in the same visual context, their expectations for VLM assistants may vary widely.
Existing general alignment objectives struggle to meet such a wide range of needs. Our research aims to bridge this gap by exploring how to align VLM assistants with personalized situated cognition.


\paragraph{Personalized Alignment of AI Models}
There are two main orientations~\cite{wang-2024-essence} of AI models' personalized alignment~\cite{kirk-2024-benefits,kirk-2024-prism,zhang-2024-PersonSurvey,sorensen-2024-roadmap,li-2024-personalized,wu-2024-personalized}: 1) \textit{Personal Reflection}, where AI imitates human personalities for AI-based social simulation~\cite{mou-2024-SocialSimulation,mou-2024-agentsense,zhou-2023-sotopia} or social agents~\cite{feng-2024-surveySocialAgents,agrawal-2023-multimodal,fan-2023-athena}; 2) \textit{Personal Assistant}, which tailors AI to be the ideal assistant for humans with specific personalities, including personalized recommendations~\cite{wei-2024-VLMRec}, retrieval~\cite{pi-2024-personalized,shen-2024-pmg,nguyen-2024-yollava}, or personalized image~\cite{li-2024-stylegan,shi-2024-instantbooth} or text generation~\cite{li-2024-learningRewrite} assistance. Our work falls within the scope of the latter.

The most relevant studies to our work are recent research on personalized Large Language Model (LLM) assistants~\cite{jang-2023-personalized,wu-2024-AlignInteract,sun-2024-personaDB,wang-2024-aiPERSONA}. However, these studies: 1) do not involve visual contexts; 2) mainly focus on certain combinations of personalized text styles or values~\cite{chen-2024-pad,wang-2024-map,zhang-2024-metaalign,zhang-2024-controllable,cheng-2023-everyone,lee-2024-bapo}, which are unable to model human diversity.
Our research seeks to address these issues through: 1) modeling human diversity using the sociological concept of Role-Set; and 2) investigating personalized alignment of VLM assistants in visual scene tasks.




\section{Problem Definition}

To clarify the proposed task, this section provides: 1) problem simplifications, 2) the notations, and 3) the optimization objective of this task.

\paragraph{Problem Simplification}
The proposed task aims to explore how to align VLM assistants with the personalized situated cognitions of diverse individuals.
However, there are two main challenges when exploring this problem: 1) \textit{How to define diverse individuals?} 2) \textit{How to evaluate whether the personalized alignment is achieved?}

For the first question, we introduce \textit{Role-Set} to characterize individuals' diversity. In Role Theory~\cite{goffman-2023-presentation}, a \textit{\textbf{Role-Set}} is the collection of various roles connected to specific social positions, influencing an individual's behavior and expectations. The roles within an individual's Role-Set reflect the combination of their social positions, which may lead to diverse situated cognitions and personalized expectations for VLM assistants' responses. In this study, each individual's Role-Set is set to contain 5 ``\textit{Role@Location}'' components.

For the second question, we believe that personalized alignment should enable the VLM assistant to generate personalized responses that effectively guide individuals toward taking better actions. As emphasized in~\citet{gallagher-2006-body}, both the human body and mind are critical to human action. Thus, we conceptualize ``\textit{\textbf{action}}'' here as the combination of external \textit{Body Behavior} and internal \textit{Mind Feelings}. For example, in Figure~\ref{fig:introduction_figure}, a personalized response should help the ``child'' move away from the broken swing and reduce feelings of worry.

\paragraph{Notations}
Consider a sample \( s = (RS, v, q) \), where \( RS \) denotes the Role-Set of the individual, \( v \) represents the visual scene (image) encountered by the individual, \( q \) denotes the query posed by the individual.
The VLM assistant, parameterized by \(\theta\), is represented as \( f_{\theta} \). For a given sample \( s \), the response generated by the VLM assistant is \( r = f_{\theta}(s) \).
The individual with the Role-Set $RS$ who encounters the visual scene $v$ has the situated cognition $c= C(s)$.
Subsequently, based on the individual's situated cognition $c$ and the received VLM response $r$, the probability that the individual takes the action \( a \) is denoted as $P_{A}(a|r,c)$. 



\paragraph{Optimization Objective}
According to the problem simplification, when the VLM response is well-aligned with the individual's personalized situated cognition, the individual is most likely to take the desired optimal action.
Thus, based on the above notations, we define the optimization objective as:
\begin{equation}\label{eq:optimize_objective}
    \theta^* = \argmax_{{\theta}} \mathbb{E}_{s \sim S_{\text{train}}} P_{A}(a^*|r=f_{\theta}(s),c),
\end{equation}

where $S_{\text{train}}$ is the training set, $c=C(s)$, and $s = (RS,v,q)$. 
The objective is to find the optimal parameters \(\theta^*\) for the VLM assistant that maximizes the expected probability of the individual's action being the desired action \( a^* \).



\begin{figure*}[t!]
\centering
\includegraphics[width=0.84\textwidth]{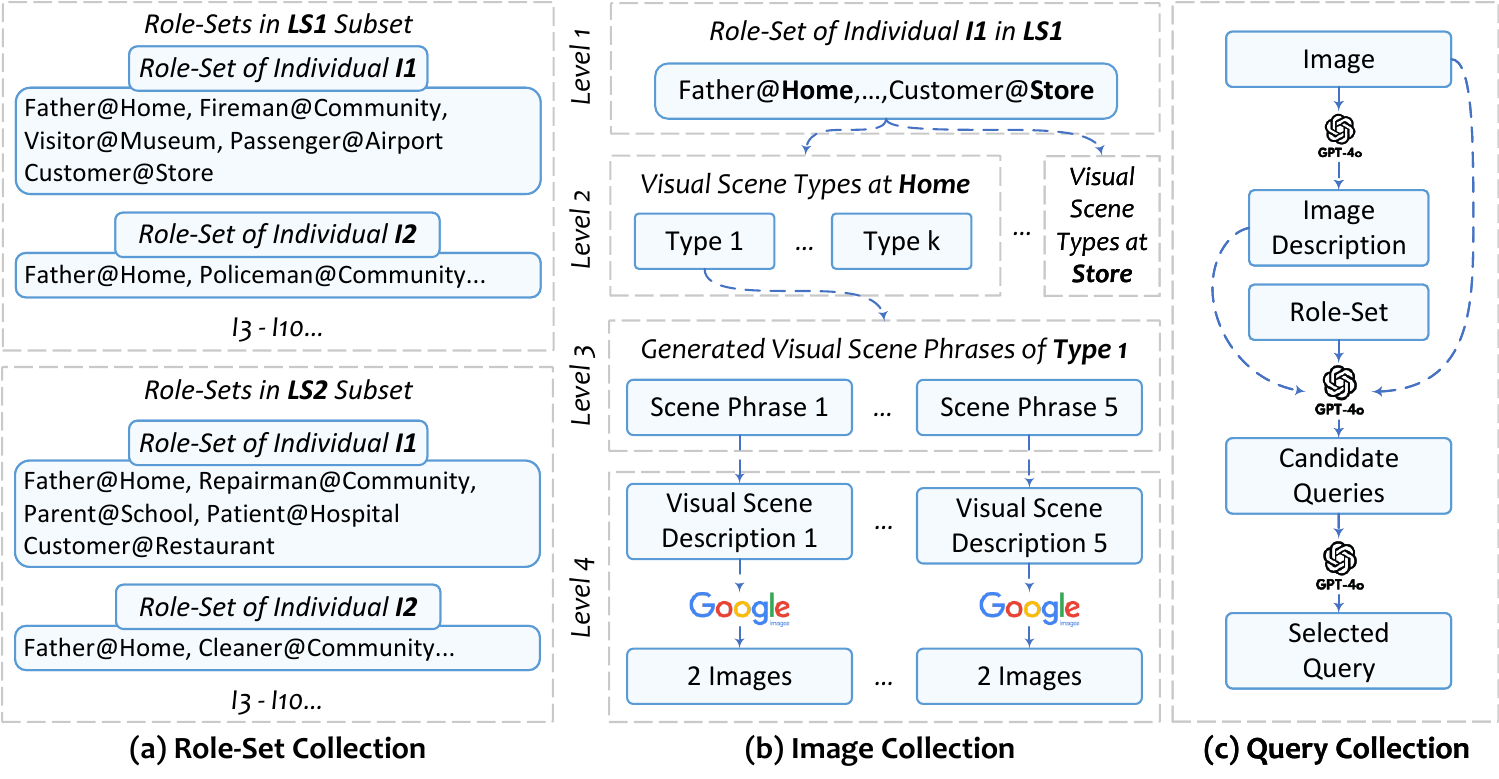}
\caption{Overview of the collection process of PCogAlignBench. 
The \textbf{LS1} subset considers roles in ``\{Home, Community, Museum, Airport, Store\}''.
The \textbf{LS2} subset considers roles in ``\{Home, Community, School, Hospital, Restaurant\}''.
Here we take the ``{Individual I1 in LS1}'' as an example to illustrate the process of image collection.}
\label{fig:benchmark_collection}
\end{figure*}

\section{PCogAlignBench}
In this section, we introduce the benchmark construction and the automatic evaluation method.
\paragraph{Benchmark Overview}
The proposed benchmark consists of 12k training samples and 6k test samples.
We define 20 individuals with different Role-Sets in total.
Each training sample consists of an individual's Role-Set, an image, and a query proposed by the individual.
To ensure the evaluation is reliable, we further gather ``\textit{oracle guidance}'' for each test sample, which describes the characteristics of the expected personalized response.

Besides, considering that the proposed benchmark cannot cover all possible Role-Sets, we need to ensure settings where the Role-Sets encountered during the training and test phases do not overlap.
To achieve this, we divide the defined 20 Role-Sets into two subsets, LS1 and LS2, according to the different social location sets considered, as shown in Figure~\ref{fig:benchmark_collection} (a).
Such division allows us to test the model's generalization ability to unseen Role-Sets, e.g., training on LS1 and testing on LS2.


\subsection{Benchmark Collection}
Figure~\ref{fig:benchmark_collection} illustrates the construction of PCogAlignBench, consisting of three steps, i.e., Role-Set collection, image collection, and query collection.

\paragraph{Role-Set Collection}
We refer to~\citet{oldenburg-1989-great} and select eight social locations where individuals might participate, including \textit{Home, Community, Museum, Airport, Store, School, Hospital, Restaurant}. 
Then, we define 3 to 5 roles for each location (32 roles in total), e.g., \textit{Student, Teacher, Librarian} in the \textit{School}.
Finally, based on certain constraints~\footnote{For example, each individual can only have one permanent role. For more details, please refer to Appendix~\ref{sec:app:role_set_collect}.}, we construct 20 different Role-Sets via combinations, 10 per subset as in Figure~\ref{fig:benchmark_collection} (a).


\paragraph{Image Collection}
To ensure the diversity and reliability of collected images via search engines, we need to first collect diverse and detailed visual scene descriptions as search terms.
To achieve this, we design a hierarchical collection strategy as shown in Figure~\ref{fig:benchmark_collection} (b).

Specifically, the collection process consists of four coarse-to-fine levels, beginning with the individual's Role-Set at \textit{Level 1}. 
At \textit{Level 2}, we collect various visual scene types that individuals might observe in the social locations associated with their Role-Sets.
However, we find that it is quite difficult to generate diverse visual scene descriptions directly through the collected types.
Thus, at \textit{Level 3}, we include a middle process, i.e., visual scene phrases collection, which can ensure diversity more easily. 
Then, at \textit{Level 4}, these collected visual scene phrases serve as seeds to guide the generation of detailed visual scene descriptions. Finally, we use these gathered visual scene descriptions as search terms in search engines to retrieve the desired images.
Throughout this process, all generation is accomplished by providing instructions and demonstrations to GPT-4o.
Please refer to Table~\ref{tab:app:prompt_SceneTypesPhraseDesc} in the appendix for the prompt templates used for collecting visual scene types, visual scene phrases, and visual scene descriptions.

\paragraph{Query Collection}
We further collect queries that individuals might pose to the VLM assistant in various image contexts. However, we find that GPT-4o may sometimes generate hallucinated queries that cannot be answered based on the information in the image. To address this, we first generate several candidate queries and then ask GPT-4o to select the best query according to predefined rules, e.g., the rule of avoiding queries that require real-time information, as illustrated in Figure~\ref{fig:benchmark_collection} (c).
Please refer to Table~\ref{tab:app:prompt_CollectQuery} in the appendix for prompt templates used for describing the collected images, generating candidate queries, and selecting the best query.


\paragraph{Quality Control}
Considering that the process of image and query collection is automated through GPT-4o and search engines, it's unavoidable that some noise data might be collected. 
To ensure the evaluation reliably reflects the model's personalized alignment performance, we perform quality control on the test split by human annotators~\footnote{In the practice of personalized alignment, training samples are difficult to gather manually and typically contain noise, so we did not apply quality control to the training split.}. 
Specifically, in the image collection, two human annotators revise about 30 duplicate or unclear visual scene types at level 2. 
At level 4 of image collection, we replace approximately 1,000 low-quality images with manually gathered ones collected by 10 trained annotators. 
For the query collection, one annotator carefully revises the queries that are observed to be unanswerable in experiments.


\subsection{Evaluation on PCogAlignBench}\label{sec:bench_eval_method}
To evaluate the personalized response automatically, we adopt the ``\textit{llm-as-a-judge}'' manner. 
We also have made two specific designs inspired by~\citet{gu-2024-JudgeSurvey} to ensure reliability.
1) We collect ``\textit{oracle guidance}'' for each test sample through a collaboration between GPT-4o and humans before evaluation, describing ``what kind of personalized response is expected by the individual''.
2) We break down the scoring criteria into five dimensions, three reflecting personalized quality based on our problem definition and two reflecting general quality.

In the evaluation process, we employ a simulated interview method where the evaluator (GPT-4o) acts as an interviewee with a specific Role-Set. The “\textit{oracle guidance}” is provided to the evaluator to score the VLM assistant's responses based on the following five dimensions~\footnote{Please refer to the Appendix~\ref{sec:app:eval_method} for detailed collection process of ``oracle guidance'' and descriptions of dimensions.}.
1) \textit{Role-Set Awareness} (\textit{\textbf{RSA}}): Advice tailored to roles.
2) \textit{Body Behavior Awareness} (\textit{\textbf{BBA}}): Strategies for user's desired behavior and physical goals support.
3) \textit{Mind Feelings Awareness} (\textit{\textbf{MFA}}): Support for user's emotional needs and desired mental states.
4) \textit{Contextual Awareness} (\textit{\textbf{CA}}): Relevance to the query and context.
5) \textit{Conversational Flow} (\textit{\textbf{CF}}): Interaction, natural flow, and balance between detail and clarity.

Each scoring dimension ranges from integer 1 to 5. 
The average score across five dimensions is referred to as the personalization score (\textbf{\textit{P. Score}}).
Besides, we conduct a human evaluation to ensure this automatic evaluation is reliable and consistent with humans, as shown in Figure~\ref{fig:heatmaps_percent}.

\section{PCogAlign}
\paragraph{Overview}
The optimization objective in Eq.~\ref{eq:optimize_objective} can be streamlined into a three-step process. 1) \textit{Estimate the variables} \(c\) and \(a^*\), which are independent of the VLM's parameters $\theta$. 2) \textit{Sample personalized responses} that may be expected by the individual. 3) \textit{Select the optimal response} \(r^*\) that maximizes the probability \(P_{A}(a^*|r^*, c)\) of the individual taking the optimal action \(a^*\). Then, the VLM parameters, \(\theta\), are optimized to maximize the likelihood that \(f_{\theta}(s) = r^*\), yielding the optimal $\theta^*$.

\begin{figure}[t!]
\centering
\includegraphics[width=0.42\textwidth]{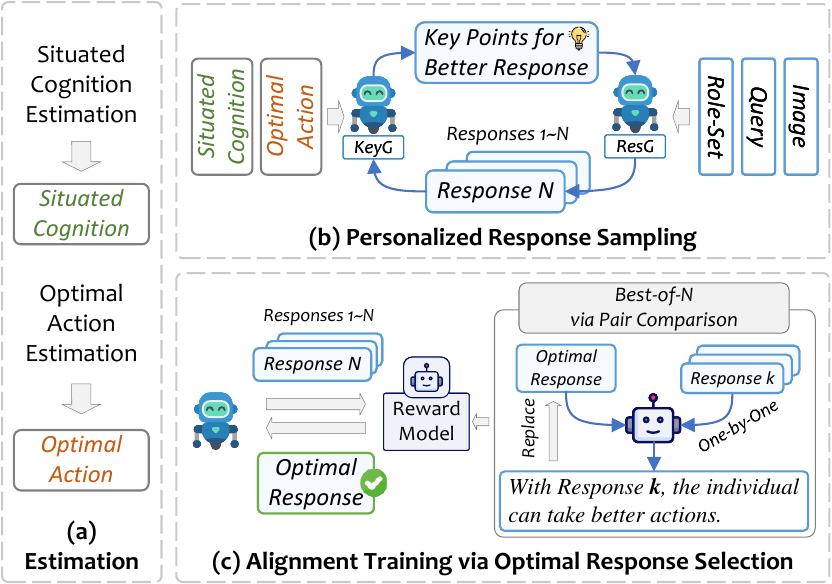}
\caption{Overview of the proposed PCogAlign. We first estimate the situated cognition and optimal action in (a). Then, the ``\textit{KeyPoints Generator (KeyG)}'' and ``\textit{Response Generator (ResG)}'' agents cooperate to sample $N$ candidate personalized responses in (b). Finally, we utilize a specific reward model to select the optimal response as feedback for alignment training in (c).}
\label{fig:method_overview}
\end{figure}

According to this, we propose a framework called PCogAlign (Figure~\ref{fig:method_overview}).
In (a), we prompt VLM to generate estimations of the situated cognition \(c\) and optimal action \(a^*\). 
In (b), we propose an iterative sampling strategy to collect personalized responses.
In (c), we construct a cognition-aware and action-based reward model, which is utilized to select the optimal response \(r^*\) as feedback to the VLM assistant for alignment training.
This section will introduce the above three steps.

\subsection{Cognition and Action Estimation}~\label{sec:method_estimation}
As shown in Figure~\ref{fig:method_overview} (a), we utilize the VLM itself to estimate the situated cognition of the individual under certain visual scenes, as well as the possible optimal action that the individual can take with certain optimal VLM assistants' responses.
This step is achieved through in-context learning using human-written demonstrations.

\subsection{Personalized Response Sampling}~\label{sec:sampling_key_points}
To collect personalized responses that may be expected by the individual, we implement an iterative sampling strategy, as shown in Figure~\ref{fig:method_overview} (b). 

Specifically, we design two cooperative agents: ``KeyG'' and ``ResG''. The ``\textit{KeyG}'' is equipped with information about the user’s situated cognition and expected optimal action, and it generates key points about how to consider the user's cognition and enhance the user's body behavior and mind feelings.
The ``\textit{ResG}'' uses the key points to re-generate responses. Through several iterations, we can gather $N$ candidate personalized responses.

\begin{figure}[t!]
\centering
\includegraphics[width=0.42\textwidth]{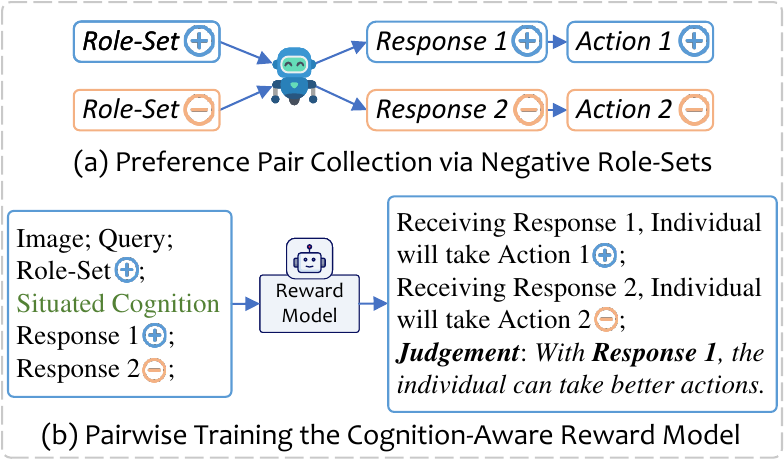}
\caption{Overview of the reward model construction.}
\label{fig:reward_model_fig}
\end{figure}

\subsection{Alignment Training}~\label{sec:alignment_training}
Since not all sampled responses align well with personalized situated cognition, we need to select the optimal one for alignment training. To facilitate this selection, we first construct a cognition-aware and action-based reward model, as depicted in Figure~\ref{fig:reward_model_fig}. 
We then use this reward model for optimal response selection in Figure~\ref{fig:method_overview} (c).

\paragraph{Reward Model Construction}
The reward model is a key component in alignment techniques like RLHF~\cite{bai-2022-AnthropicHH}, and is often trained using manually collected preference data. 
However, in personalized alignment, we cannot use such general preference data. 
Instead, we need to collect preference samples for each individual. 

To this end, we propose using \textit{negative Role-Sets} to collect preference data for reward model training as shown in Figure~\ref{fig:reward_model_fig} (a). 
For example, for individual I1 (``\textit{Teacher@School}'') and I2 (``\textit{Student@School}''), VLM's responses for I2 are inappropriate for I1.
Thus, the Role-Set of I2 can be treated as a negative one for I1.
Besides, since whether a response meets personalized expectations is reflected in the individual's actions, we incorporate the individual's actions into the preference pair.
The collected preference pairs are then used to train a cognition-aware and action-based reward model, as shown in Figure~\ref{fig:reward_model_fig} (b).

\paragraph{Optimal Response Selection via Best-of-N}
We employ a \textit{Best-of-N} strategy to select the optimal response, as shown in Figure~\ref{fig:method_overview} (c).
Specifically, each personalized response is compared with the current optimal one.
The optimal response is replaced with the new one if the reward model judges that the new one is better aligned with the personalized situated cognition and can guide the individual towards better action.

\paragraph{Optimization}
The finally selected optimal response \(r^*\) is used to optimize the VLM assistant \(f_\theta\) using the following supervised fine-tuning loss.
\begin{equation}\label{eq:sft_loss}
\theta^* = \argmin_{{\theta}} \mathbb{E}_{s \sim S_{\text{train}}} - \log P(f_\theta(r^* | s)),
\end{equation}
where the obtained \(\theta^*\) is also the solution to Eq.~\ref{eq:optimize_objective}.

\section{Experiments}

\subsection{Human Evaluation on PCogAlignBench}
In Section~\ref{sec:bench_eval_method}, we propose an automatic method to evaluate the personalization of the VLM's response.
Although the \textit{LLM-as-a-judge} manner has been validated to be largely consistent with human evaluators in general preference~\cite{zheng-2023-judging}, e.g., harmless, we still need to ensure that it is also reliable for our personalized alignment task.

\begin{figure}[t!]
\centering
\includegraphics[width=0.43\textwidth]{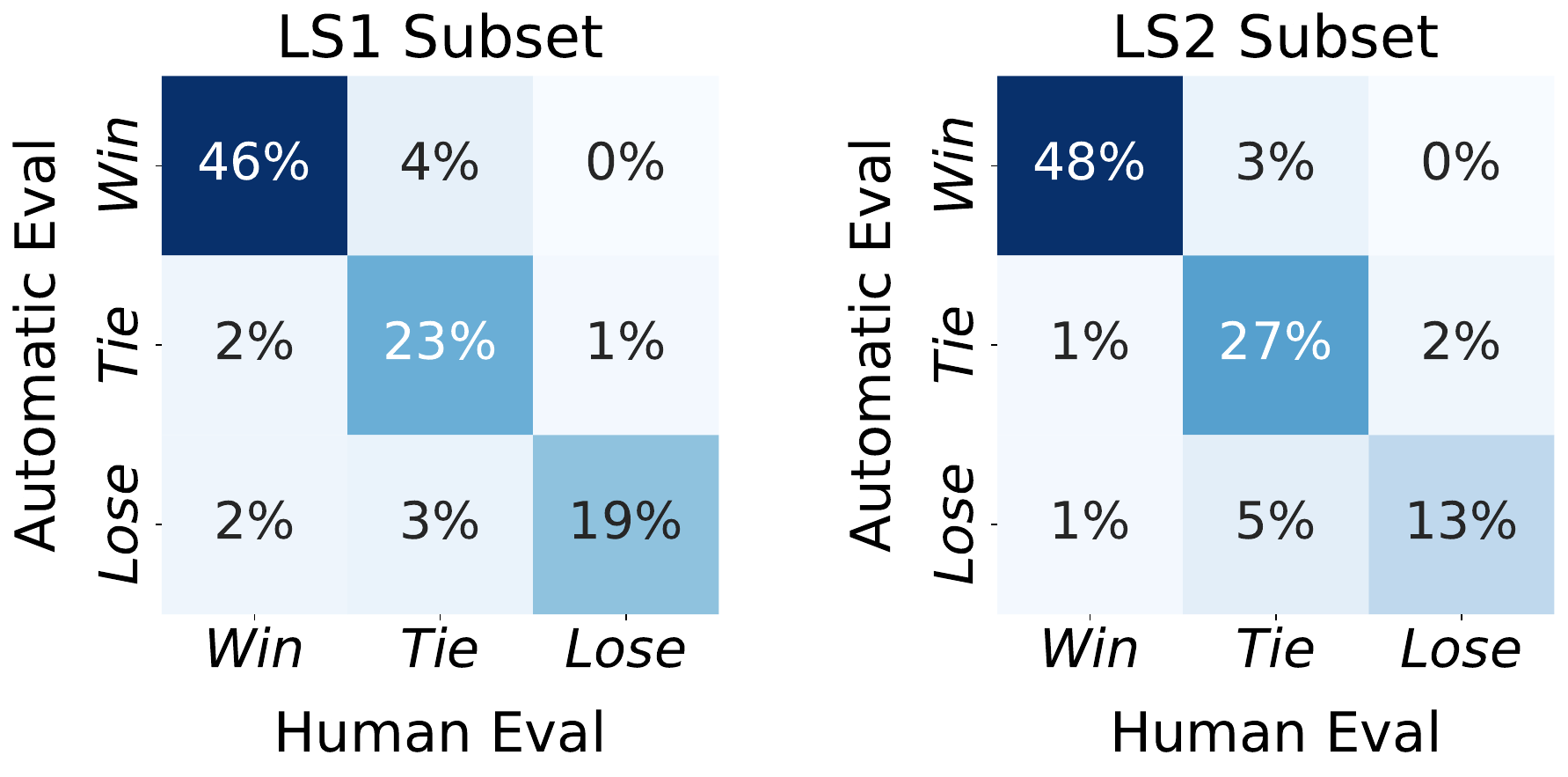}
\caption{Heatmaps showing the comparison between automatic and human evaluations on two subsets, LS1 and LS2. The automatic evaluation agrees with human assessment in 88\% of cases (the sum of the diagonal elements), highlighting the reliability of the proposed automatic evaluation method.}
\label{fig:heatmaps_percent}
\end{figure}

To this end, we randomly select 100 samples from the test split of LS1 and LS2, respectively.
Then we compare the \textit{P. Score} of responses generated by two methods, i.e., our PCogAlign and the \textit{RS Prompt} introduced below.
This yields the win/tie/lose results of the ``\textit{Automatic Eval}'' in Figure~\ref{fig:heatmaps_percent}. 
Additionally, we ask human annotators to reassess which method's response is better, which yields the results of ``\textit{Human Eval}'' in Figure~\ref{fig:heatmaps_percent}.
From the results, we can see that human evaluators agree with our designed automatic evaluation method in 88\% of the cases. This validates the reliability of the evaluation method.

\subsection{Evaluation on PCogAlign}
\subsubsection{Experimental Setup}
\paragraph{Settings}
The proposed benchmark consists of two subsets based on the roles' locations: LS1 and LS2, which facilitates the examination of methods' performance where the Role-Sets are shared or non-overlap in training and test phases.
Specifically, we consider the following settings: \textit{LS1→LS1}, \textit{LS1→LS2}, \textit{LS2→LS1}, and \textit{LS2→LS2}. For example, \textit{LS1→LS2} refers to training on LS1's train split and test on LS2's test split.


\paragraph{Used Model}
In the main experiments, both the VLM to be optimized and the cognition-aware reward model are initialized using the Qwen2-VL-7B-Instruct~\cite{wang-2024-qwen2vl}.

\paragraph{Evaluation Metrics}
Based on the evaluation method described in Section~\ref{sec:bench_eval_method}, we report two metrics in the main results. (1) \textbf{P. Score}, which ranges from 1 to 5. (2) \textbf{Win Rate}, which is obtained by comparing the \textit{P. Score} of the response being evaluated with that of the \textit{Base} response.


\subsubsection{Baselines}
Since the proposed task is new, we can only modify existing alignment methods as baselines.
We consider three types of baselines: prompt, direct preference optimization (DPO)~\cite{rafailov-2024-direct}, and supervised fine-tuning (SFT)~\cite{wei-2021-finetunedSFT}.

To ensure a fair comparison, we empower all the DPO and SFT baselines with the estimated situated cognition and optimal action in Section~\ref{sec:method_estimation} and generated ``\textit{Key Points}'' in Section~\ref{sec:sampling_key_points}~\footnote{Please refer to Appendix~\ref{sec:app:experimental_details} for more experimental details.}.

\paragraph{Prompt Baselines}
We include three prompt-based baselines, i.e., \textit{Base}, \textit{RAG}, and \textit{RS Prompt}.

\textit{Base} generates responses via the initial VLM.

\textit{RS Prompt}, where the Role-Set is provided in the system prompt to generate responses.

\textit{RAG} retrieves 3 in-context examples from training samples based on Role-Sets. We use the Levenshtein distance as the retrieval metric because it can accurately reflect the similarity between Role-Sets.

\paragraph{DPO (D) Baselines}
We include the following methods to collect preference pairs used for DPO.

\textit{Self-Refine (D)}~\cite{ranaldi-2024-self,madaan-2024-self}, which designs 3 agents to iteratively improve the generated personalized responses. We treat the finally obtained response as the chosen response and the response via \textit{RS Prompt} as the rejected one.

\textit{RLCD (D)}~\cite{yang-2023-rlcd}, which is adapted to use the \textit{Response Generator} in Section \ref{sec:sampling_key_points} to generate chosen responses. The rejected one is also obtained via \textit{RS Prompt}.

\textit{RLAIF (D)}~\cite{lee-2023-rlaif,yu-2024-rlaifv}, which utilizes a \textit{Judge Agent}, to determine the chosen and rejected response from the two responses obtained in \textit{RLCD (D)}.

\paragraph{SFT (S) Baselines}
We include the following methods to collect personalized responses used for SFT.
For the \textit{RS Prompt (S)}, we use the \textit{RS Prompt} described above to generate personalized responses as the SFT targets.
For \textit{Self-Refine (S)}, \textit{RLCD (S)}, and \textit{RLAIF (S)}, we adopt the chosen response in the DPO versions as the target for SFT.

\paragraph{PCogAlign Variants}
We have also implemented the Prompt/DPO/SFT variant of our proposed framework as ablation variants, i.e., PCogAlign (P/D/S).
The PCogAlign (P) variant only retains cooperative agents in Section \ref{sec:method_estimation} and \ref{sec:sampling_key_points} to generate responses for the test samples.
The PCogAlign (D) variant adopts the reward model to conduct multiple pairwise comparisons with the response via \textit{RS Prompt}.
For the PCogAlign (S), we remove the Best-of-N strategy and take the first response chosen by PCogAlign (D) that is not the response of \textit{RS Prompt} for each sample as the SFT target.

\begin{table*}[]

    \resizebox{\linewidth}{!}
    {
\begin{tabular}{clcccccccccc}
        \toprule

\multicolumn{2}{c}{\multirow{2}{*}{\textbf{Method}}} & \multicolumn{2}{c}{\textbf{LS1→LS1}} & \multicolumn{2}{c}{\textbf{LS1→LS2}} & \multicolumn{2}{c}{\textbf{LS2→LS1}} & \multicolumn{2}{c}{\textbf{LS2→LS2}} & \multicolumn{2}{c}{\textbf{Average}}  \\
\cmidrule(lr){3-4}\cmidrule(lr){5-6}\cmidrule(lr){7-8}\cmidrule(lr){9-10}\cmidrule(lr){11-12}
\multicolumn{2}{c}{} & \textit{\footnotesize P. Score}   & \textit{\footnotesize Win Rate}  & \textit{\footnotesize P. Score}  & \textit{\footnotesize Win Rate} & \textit{\footnotesize P. Score}  & \textit{\footnotesize Win Rate} & \textit{\footnotesize P. Score}  & \textit{\footnotesize Win Rate} & \textit{\footnotesize P. Score} & \textit{\footnotesize Win Rate} \\

\midrule

\multirow{3}{*}{\rotatebox{90}{\textit{Prompt}}}   & Base          &3.781    &0.0\%   &3.715    &0.0\%   &3.781    &0.0\%   &3.715    &0.0\%   &3.748  &0.0\%             \\
               &RS Prompt     &3.989    &44.1\%  &4.008    &47.1\%  &3.989    &44.1\%  &4.008    &47.1\%  &3.999  &45.6\%            \\
               &RAG     &4.022 & 43.9\% & 3.949 & 44.3\% & 4.046 & 46.4\% & 3.952 & 45.8\% & 3.992 & 45.1\%         \\
               &PCogAlign (P)   & 4.070    &47.5\%  &4.056    &50.3\%  &4.070    &47.5\%  &4.056    &50.3\%  &4.063  &48.9\%            \\
\midrule
\multirow{4}{*}{\rotatebox{90}{\textit{DPO}}}      & Self-Refine (D)          & 3.982    &43.7\%  &4.007    &47.2\%  &3.990    &43.9\%  &4.005    &46.4\%  &3.996  &45.3\%            \\
               &RLCD (D)      &3.985    &44.6\%  &4.014    &46.7\%  &3.978    &43.2\%  &4.006    &47.1\%  &3.996  &45.4\%            \\
               &RLAIF (D)     &3.984    &43.6\%  &4.005    &46.9\%  &3.982    &43.8\%  &4.011    &48.4\%  &3.996  &45.7\%            \\
               &PCogAlign (D)   & 4.020    &45.2\%  &4.039    &49.1\%  &4.008    &45.2\%  &4.029    &48.1\%  &4.024  &46.9\%            \\
\midrule
\multirow{5}{*}{\rotatebox{90}{\textit{SFT}}}      & RS Prompt (S) &3.857    &35.3\%  &3.937    &41.6\%  &3.934    &38.7\%  &3.969    &42.5\%  &3.924  &39.5\%            \\
               &Self-Refine (S)          & 4.091    &50.3\%  &4.096    &52.0\%  &\underline{4.126}    &\underline{51.1\%}  &\underline{4.139}    &\underline{52.4\%}  &{4.113}  &\underline{51.4}\%            \\
               &RLCD (S)        & 3.996    &43.2\%  &3.941    &44.4\%  &4.016    &42.7\%  &3.965    &44.3\%  &3.980  &43.7\%            \\
               &RLAIF (S)     &4.046    &45.1\%  &3.994    &48.7\%  &4.050    &42.3\%  &4.043    &46.5\%  &4.033  &45.6\%            \\
               &PCogAlign (S)     &\underline{4.130}    &\underline{51.5\%}  &\underline{4.108}    &\underline{53.0\%}  &{4.116}    &{48.7\%}  &{4.118}    &{51.9\%}  &\underline{4.118}  &{51.3\%}            \\

\midrule
\multicolumn{2}{c}{\textbf{PCogAlign}}    &\textbf{4.161}    &\textbf{53.3\%}  &\textbf{4.156}    &\textbf{56.6\%}  &\textbf{4.150}    &\textbf{51.4\%}  &\textbf{4.151}    &\textbf{53.8\%}  &\textbf{4.154}  &\textbf{53.8\%} \\

\bottomrule

\end{tabular}
    }
\caption{Experimental results (\textit{P. Score} and \textit{Win Rate}) under four settings. The best results are in \textbf{bold} and the second best ones are \underline{underlined}.
Please refer to Table~\ref{tab:app:detailed_WinTieLose} for a comprehensive report on the \textit{Win/Tie/Lose Rate} metrics.
}
\label{tab:main_results}
\end{table*}

\subsubsection{Results and Analysis}\label{sec:main_results}
Table~\ref{tab:main_results} reports the experimental results of baselines and our proposed PCogAlign. Based on these results, we have made the following observations.

1) Our method achieves the best performance, with an average improvement of 2.4\% in \textit{Win Rate} compared to the second best \textit{Self-Refine (S)}.

2) The \textit{P. Score} of the \textit{Base} method can only reach 3.748, indicating that the VLM's general responses struggle to meet the diverse expectations brought by the personalized situated cognition.

3) Even with the simple prompt-based variant of our framework, i.e., \textit{PCogAlign (P)}, a 3.3\% improvement in \textit{Win Rate} compared to the \textit{RS Prompt} can be achieved. This indicates that the cooperative agents designed for personalized response sampling can gather expected responses that align with the personalized situated cognition.

4) An alternative approach to the reward model used for preference judgment is designing a judge agent based on VLM itself, as \textit{RLAIF (D/S)} does. The experiments show that our ablated version \textit{PCogAlign (D/S)} achieves up to a 5.7\% improvement over \textit{RLAIF (D/S)}, indicating that our proposed cognition-aware and action-based reward model effectively empowers the optimal personalized response selection.


\begin{table}[]
    \resizebox{\linewidth}{!}
    {
\begin{tabular}{clcccccccccc}
        \toprule

\multicolumn{2}{c}{\textbf{Method}}       & \multicolumn{1}{c}{\textit{\textbf{RSA}}} & \multicolumn{1}{c}{\textit{\textbf{BBA}}} & \multicolumn{1}{c}{\textit{\textbf{MFA}}} & \multicolumn{1}{c}{\textit{\textbf{CA}}} & \multicolumn{1}{c}{\textit{\textbf{CF}}} \\
\midrule
\multirow{3}{*}{\rotatebox{90}{\textit{Prompt}}} & Base            & 3.542 &3.511 &3.575 &4.447 &3.665      \\
                        & RS Prompt       & 3.842 &3.724 &3.869 &4.711 &3.846      \\
                        & RAG             & 3.805 & 3.741 & 3.853 & 4.691 & 3.870    \\
                        & PCogAlign (P)   & 3.863 &3.758 &3.978 &4.765 &3.951      \\
\midrule
\multirow{4}{*}{\rotatebox{90}{\textit{DPO}}}    & Self-Refine (D) & 3.838 &3.722 &3.864 &4.709 &3.846      \\
                        & RLCD (D)        & 3.834 &3.728 &3.865 &4.705 &3.847      \\
                        & RLAIF (D)       & 3.833 &3.739 &3.858 &4.705 &3.843      \\
                        & PCogAlign (D)   & 3.865 &3.752 &3.899 &4.735 &3.869      \\
\midrule
\multirow{5}{*}{\rotatebox{90}{\textit{SFT}}}    & RS Prompt (S)   & 3.749 &3.636 &3.748 &4.663 &3.826      \\
                        & Self-Refine (S) & \underline{3.930} &\underline{3.846} &\underline{4.049} &4.807 &3.932      \\
                        & RLCD (S)        & 3.757 &3.665 &3.865 &4.705 &3.905      \\
                        & RLAIF (S)       & 3.847 &3.730 &3.903 &4.763 &3.924      \\
                        & PCogAlign (S)       & 3.927 &3.819 &4.043 &\underline{4.822} &\underline{3.978}      \\

\midrule

\multicolumn{2}{c}{\textbf{PCogAlign}}    & \textbf{3.963}                            & \textbf{3.875}                            & \textbf{4.103}                            & \textbf{4.838}                           & \textbf{3.993}                    \\

\bottomrule

\end{tabular}
    }
\caption{Experimental results on the 5 detailed dimensions. The score ranges from 1 to 5. The best results are in \textbf{bold} and the second best ones are \underline{underlined}.}
\label{tab:results_dimensions}
\end{table}


\paragraph{Analysis on Detailed Dimensions}
Table 2 reports the results on the 5 dimensions described in Section~\ref{sec:bench_eval_method}, which is averaged by all settings in Table~\ref{tab:main_results}.
We make the following observations.

First, on dimensions focused on personalization, i.e., \textit{RSA, BBA, and MFA}, our PCogAlign shows consistent improvements. This indicates that the VLM assistant obtained through PCogAlign can effectively recognize the user's Role-Set (\textit{RSA}), guide the user towards better body behavior (\textit{BBA}), and enhance mind feelings (\textit{MFA}).

Second, our PCogAlign also brings improvements in the dimensions for general quality, i.e., \textit{CA and CF}.
This suggests that our method can improve the contextual awareness (\textit{CA}) and conversational flow (\textit{CF}) of the responses.




\begin{table}[]
    \resizebox{\linewidth}{!}
    {
\begin{tabular}{lcccccc}
\toprule
\multicolumn{1}{c}{}                            & \multicolumn{3}{c}{\textbf{LS1}}     & \multicolumn{3}{c}{\textbf{LS2}}   \\
\cmidrule(lr){2-4}\cmidrule(lr){5-7}
\multicolumn{1}{c}{\multirow{-2}{*}{\textbf{}}} & \textit{hit@1} & \textit{hit@2} & \textit{hit@3} & \textit{hit@1} & \textit{hit@2} & \textit{hit@3} \\
\midrule
w/o RM                                      & 28\% & 56\%  & 69\%   &  31\% &   51\%    & 68\%      \\
w/ RM                                       & 69\%  & 94\% & 99\%   &  79\% & 93\% &   98\%  \\
\bottomrule
\end{tabular}
    }
\caption{Human evaluation results (\textit{hit@k}) of the selected responses without and with the constructed reward model (RM). The \textit{hit@k} indicates if the response is within the top $k$ of the $N$ candidates.}
\label{tab:results_hitk}
\end{table}

\begin{table*}[t]

    \resizebox{\linewidth}{!}
    {
\begin{tabular}{clcccccccccc}
        \toprule

\multicolumn{2}{c}{\multirow{2}{*}{\textbf{Method}}} & \multicolumn{2}{c}{\textbf{LS1→LS1}} & \multicolumn{2}{c}{\textbf{LS1→LS2}} & \multicolumn{2}{c}{\textbf{LS2→LS1}} & \multicolumn{2}{c}{\textbf{LS2→LS2}} & \multicolumn{2}{c}{\textbf{Average}}  \\
\cmidrule(lr){3-4}\cmidrule(lr){5-6}\cmidrule(lr){7-8}\cmidrule(lr){9-10}\cmidrule(lr){11-12}
\multicolumn{2}{c}{} & \textit{\footnotesize P. Score}   & \textit{\footnotesize Win Rate}  & \textit{\footnotesize P. Score}  & \textit{\footnotesize Win Rate} & \textit{\footnotesize P. Score}  & \textit{\footnotesize Win Rate} & \textit{\footnotesize P. Score}  & \textit{\footnotesize Win Rate} & \textit{\footnotesize P. Score} & \textit{\footnotesize Win Rate} \\

\midrule

\multirow{4}{*}{Qwen2-VL-7B-Instruct} & Base                          & 3.781 & 0.0\% & 3.715 & 0.0\% & 3.781 & 0.0\% & 3.715 & 0.0\% & 3.748 & 0.0\%             \\
                  & RS Prompt                     & 3.989 & 44.1\% & 4.008 & 47.1\% & 3.989 & 44.1\% & 4.008 & 47.1\% & 3.999 & 45.6\%            \\
                  & RAG                           & 4.022 & 43.9\% & 3.949 & 44.3\% & 4.046 & 46.4\% & 3.952 & 45.8\% & 3.992 & 45.1\%            \\
                  & \textbf{PCogAlign (P)}        & \textbf{4.070}    & \textbf{47.5\%}   & \textbf{4.056}    & \textbf{50.3\%}   & \textbf{4.070}    & \textbf{47.5\%}   & \textbf{4.056}    & \textbf{50.3\%}   & \textbf{4.063}    & \textbf{48.9\%}   \\
                  \midrule
\multirow{4}{*}{Qwen2.5-VL-7B-Instruct}  & Base                          & 4.126 & 0.0\% & 4.079 & 0.0\% & 4.126 & 0.0\% & 4.079 & 0.0\% & 4.102 & 0.0\%             \\
                  & RS Prompt                     & 4.163 & 31.0\% & 4.122 & 33.0\% & 4.163 & 31.0\% & 4.122 & 33.0\% & 4.143 & 32.0\%            \\
                  & RAG                           & 4.242 & 38.1\% & 4.184 & 37.5\% & 4.220 & 36.4\% & 4.200 & 37.3\% & 4.212 & 37.3\%            \\
                  & \textbf{PCogAlign (P)}        & \textbf{4.275}    & \textbf{40.8\%}   & \textbf{4.277}    & \textbf{44.2\%}   & \textbf{4.275}    & \textbf{40.8\%}   & \textbf{4.277}    & \textbf{44.2\%}   & \textbf{4.276}    & \textbf{42.5\%}   \\
                  \midrule
\multirow{4}{*}{Phi-3.5-vision-instruct} & Base                          & 3.268 & 0.0\% & 3.235 & 0.0\% & 3.268 & 0.0\% & 3.235 & 0.0\% & 3.251 & 0.0\%             \\
                  & RS Prompt                     & 3.419 & 40.3\% & 3.379 & 38.1\% & 3.419 & 40.3\% & 3.379 & 38.1\% & 3.399 & 39.2\%            \\
                  & RAG                           & 3.730 & 65.4\% & 3.613 & 58.8\% & 3.772 & 67.0\% & 3.684 & 62.1\% & 3.700 & 63.3\%            \\
                  & \textbf{PCogAlign (P)}        & \textbf{3.797}    & \textbf{67.3\%}   & \textbf{3.745}    & \textbf{64.2\%}   & \textbf{3.797}    & \textbf{67.3\%}   & \textbf{3.745}    & \textbf{64.2\%}   & \textbf{3.771}    & \textbf{65.7\%}   \\
                  \midrule
\multirow{4}{*}{MiniCPM-V-2\_6}          & Base                          & 3.743 & 0.0\% & 3.725 & 0.0\% & 3.743 & 0.0\% & 3.725 & 0.0\% & 3.734 & 0.0\%             \\
                  & RS Prompt                     & 4.055 & 60.7\% & 4.037 & 60.7\% & 4.055 & 60.7\% & 4.037 & 60.7\% & 4.046 & 60.7\%            \\
                  & RAG                           & 4.261 & 79.8\% & 4.232 & 78.4\% & 4.262 & 79.3\% & 4.235 & 78.8\% & 4.248 & 79.1\%            \\
                  & \textbf{PCogAlign (P)}        & \textbf{4.303}    & \textbf{80.1\%}   & \textbf{4.321}    & \textbf{81.0\%}   & \textbf{4.303}    & \textbf{80.1\%}   & \textbf{4.321}    & \textbf{81.0\%}   & \textbf{4.312}    & \textbf{80.5\%}   \\

\bottomrule

\end{tabular}
    }
\caption{Experimental results (\textit{P. Score} and \textit{Win Rate}) under four settings. The best results on each VLM are in \textbf{bold}.}
\label{tab:different_VLMs_results}
\end{table*}

\paragraph{Analysis on Constructed Reward Model}
To delve into the effectiveness of the constructed reward model in Section~\ref{sec:alignment_training}, we conduct a human evaluation to examine the actual quality of the selected optimal response by the reward model.

Specifically, we select 100 samples from the training processes on LS1 and LS2, respectively.
Human evaluators are asked to rank the top 3 responses among the $N$ candidates for each sample ($N=6$).
Then, we calculate the \textit{hit@k} metric of the selected response without and with the reward model.
Note that without the reward model, we treat the first response generated during the personalized response sampling as the selected one.

Table~\ref{tab:results_hitk} presents the human evaluation results. 
We can observe that the optimal response selected by the reward model has a 98.5\% chance of being within the top 3, indicating the high quality.
Besides, the constructed reward model brings a 44.5\% increase of \textit{hit@1}, which further validates the importance of utilizing the constructed reward model for selecting the optimal personalized response.



\paragraph{Evaluating Different VLMs on PCogAlignBench}
Our proposed PCogAlignBench can also serve as a benchmark for evaluating the personalization adaptation ability of different VLMs. To this end, we have included several VLMs from additional series to benchmark their personalization adaptation abilities. We consider the prompt-based baselines and the prompt variant of our PCogAlign for this experiment. The results are shown in Table~\ref{tab:different_VLMs_results}.

As shown in Table~\ref{tab:different_VLMs_results}, the \texttt{MiniCPM-V-2\_6} shows the strongest personalization adaptation ability (average \textit{P. Score}). Still, our PCogAlign (P) method consistently outperforms the baselines across all VLMs.

\section{Conclusion}
In this paper, we highlight the importance of a novel task, i.e., {aligning vision-language model (VLM) assistants with personalized situated cognition}. 
To explore this task, we construct a new benchmark named PCogAlignBench, consisting of 18k samples.
Furthermore, we present a novel framework named PCogAlign as a baseline for this problem, involving a cognition-aware and action-based reward model for alignment training. 
Experimental results and human evaluations validate the reliability of the proposed automatic evaluation method and the effectiveness of the PCogAlign.

\section*{Limitations}
We summarize the potential limitations of our study as follows, which may inspire future research to further explore based on this work.

1) For the experimental feasibility, we introduce the Role-Set concept to simplify the problem when defining diverse individuals. However, in real life, an individual's diversity may extend beyond the Role-Set and can be influenced by various factors such as personality and background. Fully capturing such diversity while ensuring experimental feasibility is a highly challenging issue, and we will continue to explore it in future work.

2) In the proposed PCogAlign framework, we employ a prompt-based method to estimate the personalized situated cognition and optimal action. Although our experiments have demonstrated the effectiveness of this simple yet effective estimation method, we believe there may be better ways to accomplish this step. We encourage future research to explore this more deeply.

3) Experimentally, we notice that the effects brought by DPO-based variants are very weak. However, given the important role of preference optimization algorithms in general alignment, we may have not used the most suitable preference optimization algorithm for personalized alignment. We encourage future work to address this challenge theoretically and experimentally.


\section*{Ethics Statement}

During the data collection process, we made necessary designs in the role definition phase and quality control process to mitigate potential ethical risks. Specifically, in the role definition phase, we thoroughly discussed and defined roles that meet ethical standards to avoid factors like gender bias. For instance, to avoid gender bias, we defined that the responsibilities of ``mother@home" and ``father@home" are fairly distributed, including shared household chores, childcare, and handling family emergencies. Besides, in our human-led quality control process, we aimed to avoid potential ethical risks, including gender and racial biases etc. For example, if an annotator repeatedly encounters images showing ``women doing housework" and relatively few ``men doing housework'' images, they are required to replace some of the ``women doing housework'' images with ``men doing housework''. Thanks to the carefully designed role definitions, which consider avoiding potential bias, the automatically generated visual scene descriptions have largely avoided such biases, and we encountered less than 1\% of such cases during the quality control process.

Considering the vast number of potential role combinations (theoretically 6300 combinations in our setup), which significantly reduces the feasibility of conducting academic experiments, we selected a subset of role combinations (20 Role-Sets) to form the Role-Sets in our dataset. This might raise concerns about certain biases. Although we believe this is acceptable in a simulated research environment, in actual industry development, we encourage companies/developers to consider various user backgrounds to form comprehensive, unbiased Role-Sets for data collection and personalized alignment training.

The ``diversity'' of the dataset encompasses two aspects: increasing the diversity of scenarios, and ensuring diversity to avoid potential biases, such as racial and gender biases.

We obtained informed consent from all annotators about the statement of ``all human annotators being paid by the laboratory following local wage requirements'' before the manuscript submission.

\section*{Acknowledgments}
This work was supported by the grant from the National Natural Science Foundation of China (NSFC) project (No. 62276193), the grant from Zhongguancun Academy (Grant No. 20240302), and the Fundamental Research Funds for the Central Universities, China (Grant No. 2042022dx0001).


\bibliography{custom}

\clearpage
\newpage

\appendix

\section{Supplementary Experimental Results}
\paragraph{Detailed Results}\label{sec:app:detailed_results_of_main}
In this subsection, we will present some detailed experimental results.

Specifically, Table~\ref{tab:app:detailed_WinTieLose} presents the detailed results of \textit{Win/Tie/Lose} metrics corresponding to Table~\ref{tab:main_results}.
Table~\ref{tab:app:detailed_dimen_sub1} and Table~\ref{tab:app:detailed_dimen_sub2} present the detailed results of various dimensions corresponding to Table~\ref{tab:results_dimensions}.

\paragraph{Experiments on Qwen2.5-VL}\label{sec:app:expe_on_qwen25}
We also select the recently released state-of-the-art VLM series model, Qwen2.5-VL~\cite{qwen2.5-VL}, to conduct scalability experiments. We choose the ``LS1→LS2'' cross-subset setting for the experiment and report the results on Qwen2.5-VL-3B-Instruct in Table~\ref{tab:app:qwen25_3b}. 
From the experimental results, we can observe conclusions consistent with those in the main text section (Section~\ref{sec:main_results}).

\section{PCogAlignBench Details}

\subsection{Role-Set Collection}\label{sec:app:role_set_collect}
We reference~\citet{oldenburg-1989-great} and select eight social locations where individuals might participate, including \textit{Home, Community, Museum, Airport, Store, School, Hospital, Restaurant}.
Then, for each location, we define 3 to 5 roles as shown in Table~\ref{tab:app:roles}.
Specifically, the \underline{underlined} roles are \textit{permanent roles}, which occupy the majority of an individual's daily working time.

Then, since we need to enable settings where the Role-Sets encountered during the training and test phases do not overlap, we divide the above locations into two subsets, LS1 and LS2, according to the different social location sets considered.
The LS1 consists of ``\{Home, Community, Museum, Airport, Store\}'' and the LS2 consists of ``\{Home, Community, School, Hospital, Restaurant\}''.

Finally, we construct 10 Role-Sets for each subset via combinations based on the following pre-defined constraints.
1) Each individual's Role-Set can contain only one permanent role. 2) Each individual's Role-Set involves five roles. 3) Each permanent role can be adopted by only one individual.
The collected Role-Sets are shown in Table~\ref{tab:app:role_sets_details}.

\subsection{Image Collection}\label{sec:app:img_collect}
The image collection in Figure~\ref{fig:benchmark_collection} involves three prompt templates used for collecting visual scene types, visual scene phrases, and visual scene descriptions via GPT-4o. We present the specific templates in Table~\ref{tab:app:prompt_SceneTypesPhraseDesc}.

Besides, for collecting images based on visual scene descriptions in level 4, we use Google's API platform~\footnote{https://console.cloud.google.com/apis} for automated collection.

\paragraph{Statistics on the Quality of the Image Collection}
Generally, approximately 1,000 images in the test samples (17\%) were replaced by human-collected ones. 
During this process, annotators not only need to ensure that the images show proper scenarios in the expected locations, but also need to consider the diversity of the images and avoid potential ethical risks, such as gender and racial biases. For example, even if an image is high-quality, if it appears multiple times (>2 times) in the dataset, it is replaced with a new one. Besides, suppose an annotator repeatedly encounters images showing ``women doing housework" and relatively few ``men doing housework" images. In that case, they need to replace some of the ``women doing housework" images with ``men doing housework". Thanks to the carefully designed role definitions, which consider avoiding potential bias, the automatically generated visual scene descriptions have largely avoided such biases, and we encountered less than 1\% of such biased cases during the quality control process.

\subsection{Query Collection}\label{sec:app:query_collect}
The query collection in Figure~\ref{fig:benchmark_collection} involves three prompt templates, which are used for describing collected images, generating several candidate queries, and selecting the best query, respectively, as shown in Table~\ref{tab:app:prompt_CollectQuery}.

\paragraph{Statistics on the Quality of the Query Collection}
Thanks to the well-designed query collection strategy and high-quality human-checked images, only about 20 queries (0.3\%) were found to be unanswerable and revised by annotators.

\subsection{Recruitment and Training Details for Annotators}

We recruited 10 annotators from the list of authors and lab members, all of whom are master or Ph.D. students with a background in natural language processing or computer vision. These annotators were informed about the specific requirements for the expected high-quality samples during a group meeting. Subsequently, the annotators reviewed a small set of automatically generated samples and replaced 20 low-quality samples based on their understanding of the quality control requirements. Following this, the benchmark construction manager, with the help of two other annotators, checked the samples replaced by the annotators. Finally, the benchmark construction manager provided feedback to the annotators to help them better comprehend the quality control requirements.

\begin{table}[]
    \resizebox{\linewidth}{!}
    {
\begin{tabular}{lp{8cm}}
\toprule
\textbf{Location} & \textbf{Roles} \\
\midrule
\textit{Home} & Child; Father; Mother; Grandpa; Grandma; \\
\textit{Community} & \underline{Fireman}; \underline{Policeman}; \underline{Repairman}; \underline{Cleaner}; Member; \\
\textit{Museum} & \underline{Guide}; \underline{Security Staff}; Visitor; \\
\textit{Airport} & \underline{Airline Staff}; \underline{Information Staff}; \underline{Janitor}; Passenger; \\
\textit{Store} & \underline{Cashier}; \underline{Security Personnel}; \underline{Shelf Stocker}; Customer; \\
\textit{School} & \underline{Student}; \underline{Teacher}; \underline{Librarian}; Parent; \\
\textit{Hospital} & \underline{Doctor}; \underline{Nurse}; \underline{Pharmacist}; Patient; \\
\textit{Restaurant} & \underline{Chef}; \underline{Waiter}; Customer; \\
\bottomrule
\end{tabular}

    }
\caption{The pre-defined roles in various locations. The \textit{permanent roles} are \underline{underlined}, which occupy the majority of an individual's daily working time.}
\label{tab:app:roles}
\end{table}


\begin{figure*}[t!]
\centering
\includegraphics[width=0.95\textwidth]{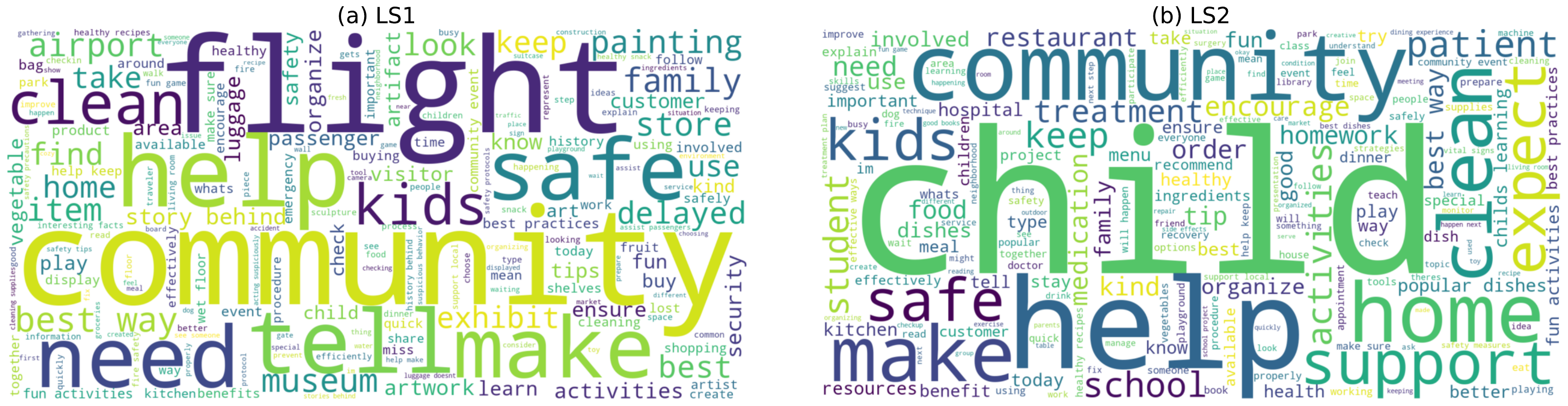}
\caption{Word cloud visualization of user queries from the test split of LS1 and LS2 subsets, illustrating the variety of queries in the dataset. Note that LS2 includes an individual who is a ``child'', and individuals may frequently encounter scenarios in various locations where child-related issues need to be addressed. This results in a significant number of queries involving ``child''.}
\label{fig:app:word_cloud_query}
\end{figure*}

\subsection{Evaluation Method}\label{sec:app:eval_method}
\paragraph{Oracle Guidance Collection}\label{sec:app:eval_method:oracle}
To collect the oracle guidance for each test sample, we adopt a collaboration manner with human annotators and GPT-4o.
Specifically, we first use GPT-4o to collect each individual's expectations in each location, resulting in 100 general expectations in total.
Then, we ask human annotators to carefully check the general expectations obtained above.
Finally, each individual's general expectation in each location is used for constructing a prompt using the prompt template in Table~\ref{tab:app:prompt_for_Oracle}, which is then used for generating the ``oracle guidance'' for each test sample.

\paragraph{Dimensions of Evaluation Criteria}\label{sec:app:eval_method:dimensions}
In Section~\ref{sec:bench_eval_method}, we propose an automatic evaluation method, which breaks down the evaluation criteria into five dimensions.
The detailed descriptions of the five dimensions are as follows:

\begin{enumerate}
    \item \textit{Role-Set Awareness} (\textit{\textbf{RSA}}): Does this response consider your multiple roles and responsibilities (especially the primary role in the specific scenario), providing advice or information specifically tailored to support you effectively? The response should provide tailored advice or information to effectively support you, acknowledging only the roles that are essential in the current context.
    \item \textit{Body Behavior Awareness} (\textit{\textbf{BBA}}): Does this response offer guidance or strategies that help you achieve your desired body behavior?
    \item \textit{Mind Feelings Awareness} (\textit{\textbf{MFA}}): Does this response provide support and address the emotional needs necessary for you to achieve your desired mind feelings?
    \item \textit{Contextual Awareness} (\textit{\textbf{CA}}): Does this response accurately address your query, maintaining focus on the main intent without deviation? Is the response relevant to your specific scenario, including location and situational factors?
    \item \textit{Conversational Flow} (\textit{\textbf{CF}}): Does this response encourage ongoing interaction by being engaging and naturally flowing? Is the response appropriately concise or detailed, delivering information that strikes a balance for optimal understanding?
\end{enumerate}

\paragraph{Prompt for Evaluation}\label{sec:app:eval_method:eval_prompt_interview}
The prompt used for evaluation is presented in Table~\ref{tab:app:EvalPrompt}.
Considering the cost, we use GPT-4o-mini for all evaluations, which, through our human evaluation, demonstrates reliable evaluation quality (Figure~\ref{fig:heatmaps_percent}).

\subsection{Word Cloud Visualization of PCogAlignBench}

To visually demonstrate the distribution of our collected dataset, we create a word cloud visualization for the queries from the collected test split, which is displayed in Figure~\ref{fig:app:word_cloud_query}. It can be observed that our benchmark includes a diverse range of visual scenes and user queries.

\section{Experimental Details}\label{sec:app:experimental_details}

\subsection{Implementation Details of PCogAlign}\label{sec:app:PCogAlign_details}
\subsubsection{Cognition and Action Estimation}\label{sec:app:PCogAlign_estimation}
As described in Section~\ref{sec:method_estimation}, given an individual's Role-Set, the visual scene (image), and the query posed by the individual, we need to first estimate the personalized situated cognition of the individual.
The prompt template used for such estimation is shown in Table~\ref{tab:app:prompt_situtated_cognition}.

Besides, we also need to estimate the possible optimal action that the individual can take after receiving the personalized response from the personalized VLM assistant.
The prompt template used for such estimation is shown in Table~\ref{tab:app:prompt_optimal_action}.

\subsubsection{Personalized Response Sampling}\label{sec:app:PCogAlign_sampling}
In Section~\ref{sec:sampling_key_points}, we design two agents, i.e., the ``\textit{KeyPoints Generator (KeyG)}'' and ``\textit{Response Generator (ResG)}'', to cooperate for sampling multiple personalized responses. The number of candidate sampled responses is set as $N=6$, including the initial response.

The prompt templates designed for the ``\textit{KeyG}'' and ``\textit{ResG}'' agents are presented in Table~\ref{tab:app:prompt_KeyG_ResG}.

\subsubsection{Reward Model Construction}\label{sec:app:PCogAlign_RM}
As shown in Figure~\ref{fig:reward_model_fig}, to construct the reward model, we first collect preference pairs used for reward model training using the negative Role-Set. 

The negative Role-Set selection is based on the different roles in certain locations.
For example, for a visual scene in ``Museum'', the currently focused positive Role-Set is ``\textit{Father@Home; Fireman@Community; \textbf{{Visitor@Museum}} Passenger@Airport; Customer@Store}''.
Then, one of the Negative Role-Set can be ``\textit{Mother@Home; Member@Community; \textbf{{Guide@Museum}}; Passenger@Airport; Customer@Store}''.

Then, we adopt the collected preference pairs to train the VLM to be a task-specific reward model via SFT.
To avoid position bias, we construct two SFT samples with different response orders for each collected preference pair.
We show an example of the SFT sample for the reward model training in Table~\ref{tab:app:template_RM_train}.

\subsubsection{Optimal Response Selection}
For the pair comparisons in the process of optimal response selection, we adopt the same prompt template as in the reward model training, i.e., Table~\ref{tab:app:template_RM_train}.

\subsubsection{Optimization}
The implementation of the SFT loss in Eq.~\ref{eq:sft_loss} follows the ``trl''~\footnote{\url{https://github.com/huggingface/trl}}.
Specifically, the learning rate is 2e-4, the batch size is 4, the learning rate scheduler type is ``cosine'', the warmup ratio is 0.03, the optimizer is ``adamw\_torch\_fused'', and the epoch is set as 4.
We also adopt LoRA~\cite{hu-2021-lora} for parameter-efficient fine-tuning where $r=8$, $\alpha=16$, and dropout is 0.05.

\subsection{Implementation Details of Baselines}\label{sec:app:baseline_details}
\paragraph{Prompt Baselines}
\begin{itemize}
    \item \textit{Base}. We directly input the query and image into the VLM to obtain the responses.
    \item \textit{RS Prompt}. Based on the \textit{Base} method, we additionally provide the Role-Set into the system prompt of VLM to obtain the responses. 
\end{itemize}

\paragraph{DPO (D) Baselines}
\begin{itemize}
    \item \textit{Self-Refine (D)}~\cite{ranaldi-2024-self,madaan-2024-self}. We design 3 agents to iteratively improve the generated personalized responses, including ``\textit{Refiner}'', ``\textit{Scorer}'', and ``\textit{Feedback Generator}''. The prompt templates used for these agents are presented in Table~\ref{tab:app:prompt_selfrefine}. The number of iterations is set to 3 because we find that additional rounds of self-refinement do not lead to an increase in scores during the self-refinement process.
    \item \textit{RLCD (D)}~\cite{yang-2023-rlcd}. We adopt the \textit{Response Generator} in Table~\ref{tab:app:prompt_KeyG_ResG} to generate the personalized responses as the chosen ones. The rejected one is also obtained via \textit{RS Prompt}.

    \item \textit{RLAIF (D)}~\cite{lee-2023-rlaif,yu-2024-rlaifv}, which utilizes a \textit{Judge Agent}, to determine the chosen and rejected response from the two responses obtained in \textit{RLCD (D)}. The prompt template designed for the \textit{Judge Agent} is presented in Table~\ref{tab:app:prompt_rlaif}.
    
\end{itemize}

\paragraph{SFT (S) Baselines}
\begin{itemize}
    \item \textit{RS Prompt (S)}. We use the \textit{RS Prompt} described above to generate personalized responses, which are used as the SFT targets.
    \item \textit{Self-Refine (S)}, \textit{RLCD (S)}, and \textit{RLAIF (S)}. We adopt the chosen response in the DPO versions as the target for SFT.
\end{itemize}

\subsection{Experimental Environment}
For all experiments, we conduct experiments on a single Nvidia A800-80G.
We use the vLLM framework~\cite{kwon-2023-vllm} for all the LLM generation.
We use the TRL framework~\cite{vonwerra-2022-trl} for all the SFT and DPO fine-tuning.


\begin{table*}[]
    \resizebox{\linewidth}{!}
    {

\begin{tabular}{clcccccccccccc}
\toprule
\multicolumn{2}{c}{\multirow{2}{*}{\textbf{Method}}} & \multicolumn{3}{c}{\textbf{LS1→LS1}}      &\multicolumn{3}{c}{\textbf{LS1→LS2}}      &\multicolumn{3}{c}{\textbf{LS2→LS1}}      &\multicolumn{3}{c}{\textbf{LS2→LS2}}                \\
\cmidrule(lr){3-5}\cmidrule(lr){6-8}\cmidrule(lr){9-11}\cmidrule(lr){12-14}
\multicolumn{2}{c}{}                       &Win↑  &Tie   &Lose↓   &Win↑  &Tie   &Lose↓  &Win↑  &Tie   &Lose↓  &Win↑  &Tie   &Lose↓            \\
\midrule
\multirow{3}{*}{\rotatebox{0}{\textit{Prompt}}}     & Base       &0.0\%   &100.0\%       & 0.0\%   &0.0\%   &100.0\%       & 0.0\%   &0.0\%   &100.0\%       & 0.0\%   &0.0\%   &100.0\%       & 0.0\%           \\
                    &RS Prompt  &44.1\%  &31.6\%  &24.2\%  &47.1\%  &32.5\%  &20.4\%  &44.1\%  &31.6\%  &24.2\%  &47.1\%  &32.5\%  &20.4\%          \\
                    &PCogAlign (P)      & 47.5\%  &26.1\%  &26.4\%  &50.3\%  &26.6\%  &23.2\%  &47.5\%  &26.1\%  &26.4\%  &50.3\%  &26.6\%  &23.2\%          \\
\midrule
\multirow{4}{*}{\rotatebox{0}{\textit{DPO}}}  &Self-Refine (D)    & 43.7\%  &32.2\%  &24.2\%  &47.2\%  &31.0\%  &21.8\%  &43.9\%  &32.4\%  &23.7\%  &46.4\%  &31.1\%  &22.5\%          \\
                    &RLCD (D)   &44.6\%  &30.1\%  &25.4\%  &46.7\%  &32.7\%  &20.6\%  &43.2\%  &31.2\%  &25.6\%  &47.1\%  &31.9\%  &21.0\%          \\
                    &RLAIF (D)  &43.6\%  &32.8\%  &23.6\%  &46.9\%  &31.7\%  &21.4\%  &43.8\%  &32.0\%  &24.2\%  &48.4\%  &30.1\%  &21.6\%          \\
                    &PCogAlign (D)      & 45.2\%  &30.7\%  &24.1\%  &49.1\%  &30.3\%  &20.6\%  &45.2\%  &31.2\%  &23.6\%  &48.1\%  &31.2\%  &20.7\%          \\
\midrule
\multirow{5}{*}{\rotatebox{0}{\textit{SFT}}}  &RS Prompt (S)      & 35.3\%  &29.6\%  &35.1\%  &41.6\%  &30.1\%  &28.2\%  &38.7\%  &29.0\%  &32.4\%  &42.5\%  &29.9\%  &27.6\%          \\
                    &Self-Refine (S)    & 50.3\%  &26.7\%  &23.1\%  &52.0\%  &27.3\%  &20.8\%  &\underline{51.1\%}  &28.0\%  &\underline{20.9\%}  &\underline{52.4\%}  &28.3\%  &\underline{19.3\%}          \\
                    &RLCD (S)   &43.2\%  &25.8\%  &31.0\%  &44.4\%  &27.6\%  &28.0\%  &42.7\%  &26.7\%  &30.6\%  &44.3\%  &27.1\%  &28.6\%          \\
                    &RLAIF (S)  &45.1\%  &28.6\%  &26.3\%  &48.7\%  &28.1\%  &23.3\%  &42.3\%  &29.8\%  &28.0\%  &46.5\%  &29.0\%  &24.5\%          \\
                    &PCogAlign (S)      & \underline{51.5\%}  &28.7\%  &\underline{19.8\%}  &\underline{53.0\%}  &28.3\%  &\underline{18.8\%}  &48.7\%  &27.4\%  &23.9\%  &51.9\%  &27.3\%  &20.8\%          \\
\midrule
\multicolumn{2}{c}{\textbf{PCogAlign}}     &\textbf{53.3\%} & {29.0\%} & \textbf{17.7\%} & \textbf{56.6\%} & {27.5\%} & {15.9\%} & \textbf{51.4\%} & {28.1\%} & \textbf{20.4\%} & \textbf{53.8\%} & {28.0\%} & \textbf{18.2\%}\\
\bottomrule
\end{tabular}

    }
\caption{Experimental results on the \textit{Win/Tie/Lose} metrics, obtained by comparing the \textit{P. Score} of methods with that of the \textit{Base} method. These results are supplementary to Table~\ref{tab:main_results} in the main text. The best results are in \textbf{bold} and the second best ones are \underline{underlined}.}
\label{tab:app:detailed_WinTieLose}
\end{table*}

\begin{table*}[]
    \resizebox{\linewidth}{!}
    {

\begin{tabular}{clcccccccccccc}
\toprule
\multicolumn{2}{c}{\multirow{2}{*}{\textbf{Method}}} & \multicolumn{6}{c}{\textbf{LS1→LS1}}  &\multicolumn{6}{c}{\textbf{LS1→LS2}}                                                                            \\
\cmidrule(lr){3-8}\cmidrule(lr){9-14}
\multicolumn{2}{c}{}      &\textit{RSA}   & \textit{BBA}   & \textit{MFA}   & \textit{CA}  &\textit{CF}  &\textit{\textbf{P. Score}} & \textit{RSA}   & \textit{BBA}   & \textit{MFA}   & \textit{CA}  &\textit{CF}  &\textit{\textbf{P. Score}} \\
\midrule
\multirow{3}{*}{\rotatebox{0}{\textit{Prompt}}}     &Base  &3.590  &3.511  &3.609  &4.505  &3.691  &3.781  &3.494  &3.510  &3.540  &4.389  &3.639  &3.715                      \\
   &RS Prompt    &3.849  &3.685  &3.843  &4.720  &3.848  &3.989  &3.836  &3.764  &3.895  &4.702  &3.843  &4.008                      \\
   &PCogAlign (P)      &3.883  &3.743  &3.962  &4.795  &3.968  &4.070  &3.843  &3.772  &3.994  &4.735  &3.933  &4.056                      \\

\midrule
\multirow{4}{*}{\rotatebox{0}{\textit{DPO}}}  &Self-Refine (D)    &3.843  &3.684  &3.832  &4.710  &3.843  &3.982  &3.831  &3.761  &3.890  &4.705  &3.850  &4.007                      \\
   &RLCD (D)     &3.845  &3.690  &3.830  &4.719  &3.844  &3.985  &3.828  &3.782  &3.911  &4.699  &3.849  &4.014                      \\
   &RLAIF (D)    &3.842  &3.695  &3.831  &4.713  &3.839  &3.984  &3.820  &3.788  &3.884  &4.696  &3.838  &4.005                      \\
   &PCogAlign (D)      &3.879  &3.712  &3.887  &4.753  &3.869  &4.020  &3.868  &3.789  &3.931  &4.729  &3.878  &4.039                      \\
\midrule
\multirow{5}{*}{\rotatebox{0}{\textit{SFT}}}  &RS Prompt (S)      &3.697  &3.543  &3.663  &4.604  &3.779  &3.857  &3.763  &3.665  &3.794  &4.644  &3.818  &3.937                      \\
   &Self-Refine (S)    &3.916  &3.802  &4.000  &4.809  &3.926  &4.091  &\underline{3.915}  &\underline{3.845}  &4.037  &4.772  &3.911  &4.096                      \\
   &RLCD (S)     &3.795  &3.660  &3.863  &4.751  &3.912  &3.996  &3.692  &3.673  &3.852  &4.613  &3.874  &3.941                      \\
   &RLAIF (S)    &3.884  &3.712  &3.903  &4.798  &3.936  &4.046  &3.785  &3.731  &3.901  &4.665  &3.887  &3.994                      \\
   &PCogAlign (S)      &\underline{3.966}  &\underline{3.812}  &\underline{4.042}  &\underline{4.846}  &\underline{3.982}  &\underline{4.130}  &3.900  &\underline{3.845}  &\underline{4.062}  &\underline{4.776}  &\underline{3.956}  &\underline{4.108}                      \\
\midrule
\multicolumn{2}{c}{\textbf{PCogAlign}}       &\textbf{3.984} & \textbf{3.861} & \textbf{4.095} & \textbf{4.857} & \textbf{4.007} & \textbf{4.161}     &\textbf{3.948} & \textbf{3.918} & \textbf{4.127} & \textbf{4.798} & \textbf{3.989} & \textbf{4.156}            \\
\bottomrule
\end{tabular}
    }
\caption{Experimental results on detailed dimensions on ``LS1→LS1'' and  ``LS1→LS2'' settings. These results are supplementary to Table~\ref{tab:results_dimensions} in the main text. The scores range from 1 to 5. The best results are in \textbf{bold} and the second best ones are \underline{underlined}.}
\label{tab:app:detailed_dimen_sub1}
\end{table*}

\begin{table*}[]
    \resizebox{\linewidth}{!}
    {

\begin{tabular}{clcccccccccccc}
\toprule
\multicolumn{2}{c}{\multirow{2}{*}{\textbf{Method}}} & \multicolumn{6}{c}{\textbf{LS2→LS1}}  &\multicolumn{6}{c}{\textbf{LS2→LS2}}                                                                            \\
\cmidrule(lr){3-8}\cmidrule(lr){9-14}
\multicolumn{2}{c}{}      &\textit{RSA}   & \textit{BBA}   & \textit{MFA}   & \textit{CA}  &\textit{CF}  &\textit{\textbf{P. Score}} & \textit{RSA}   & \textit{BBA}   & \textit{MFA}   & \textit{CA}  &\textit{CF}  &\textit{\textbf{P. Score}} \\
\midrule
\multirow{3}{*}{\rotatebox{0}{\textit{Prompt}}}     &Base  &3.590  &3.511  &3.609  &4.505  &3.691  &3.781  &3.494  &3.510  &3.540  &4.389  &3.639  &3.715                      \\
   &RS Prompt    &3.849  &3.685  &3.843  &4.720  &3.848  &3.989  &3.836  &3.764  &3.895  &4.702  &3.843  &4.008                      \\
   &PCogAlign (P)      &3.883  &3.743  &3.962  &4.795  &3.968  &4.070  &3.843  &3.772  &3.994  &4.735  &3.933  &4.056                      \\
\midrule
\multirow{4}{*}{\rotatebox{0}{\textit{DPO}}}  &Self-Refine (D)    &3.851  &3.693  &3.836  &4.719  &3.849  &3.990  &3.829  &3.751  &3.898  &4.701  &3.844  &4.005                      \\
   &RLCD (D)     &3.836  &3.674  &3.829  &4.706  &3.847  &3.978  &3.828  &3.765  &3.892  &4.695  &3.848  &4.006                      \\
   &RLAIF (D)    &3.844  &3.687  &3.829  &4.708  &3.844  &3.982  &3.827  &3.785  &3.888  &4.705  &3.851  &4.011                      \\
   &PCogAlign (D)      &3.861  &3.722  &3.860  &4.735  &3.862  &4.008  &3.854  &3.783  &3.917  &4.722  &3.867  &4.029                      \\
\midrule
\multirow{5}{*}{\rotatebox{0}{\textit{SFT}}}  &RS Prompt (S)      &3.768  &3.619  &3.729  &4.706  &3.846  &3.934  &3.769  &3.715  &3.805  &4.697  &3.860  &3.969                      \\
   &Self-Refine (S)    &\underline{3.952}  &\textbf{3.846} & \underline{4.048}  &4.839  &3.946  &\underline{4.126}  &\underline{3.938}  &\underline{3.893}  &\underline{4.111}  &\underline{4.808}  &3.944  &\underline{4.139}                      \\
   &RLCD (S)     &3.815  &3.658  &3.886  &4.781  &3.939  &4.016  &3.727  &3.667  &3.860  &4.676  &3.896  &3.965                      \\
   &RLAIF (S)    &3.883  &3.713  &3.887  &4.822  &3.947  &4.050  &3.834  &3.765  &3.921  &4.767  &3.927  &4.043                      \\
       &PCogAlign (S)      &3.942  &3.784  &4.006  &\underline{4.860}  &\underline{3.987}  &4.116  &3.899  &3.834  &4.062  &\underline{4.808}  &\textbf{3.986} & 4.118                      \\
\midrule
\multicolumn{2}{c}{\textbf{PCogAlign}}       &\textbf{3.979} & \underline{3.840} & \textbf{4.060} & \textbf{4.877} & \textbf{3.993} & \textbf{4.150}     &\textbf{3.941} & \textbf{3.881} & \textbf{4.131} & \textbf{4.819} & \underline{3.984} & \textbf{4.151}          \\
\bottomrule
\end{tabular}

    }
\caption{Experimental results on detailed dimensions on ``LS2→LS1'' and  ``LS2→LS2'' settings. These results are supplementary to Table~\ref{tab:results_dimensions} in the main text. The scores range from 1 to 5. The best results are in \textbf{bold} and the second best ones are \underline{underlined}.}
\label{tab:app:detailed_dimen_sub2}
\end{table*}

\begin{table*}[]
\footnotesize 
    \resizebox{\linewidth}{!}
    {

\begin{tabular}{clccccccccc}
\toprule
\multicolumn{2}{c}{\textbf{Method}}       & \textbf{RSA}   & \textbf{BBA}   & \textbf{MFA}   & \textbf{CA}    & \textbf{CF}    & \textbf{P. Score} & \textbf{Win↑}   & \textbf{Tie} & \textbf{Lose↓}  \\
\midrule
\multirow{3}{*}{\textit{Prompt}} & Base    &3.511  &3.556  &3.612  &4.436  &3.662  &3.755     &0.0\%   &100.0\%      & 0.0\%           \\
                &RS Prompt       & 3.953  &3.930  &4.073  &4.759  &3.885  &4.120     &57.8\%  &29.1\%       & \underline{13.1\%}          \\
                & PCogAlign (P) & 4.028 & 4.021 & 4.183 & \underline{4.854} & \textbf{3.967} & 4.210 & \underline{59.9\%} & 25.8\% & 14.3\%  \\
\midrule

\multirow{4}{*}{\textit{DPO}}    & Self-Refine (D) & 3.944  &3.936  &4.056  &4.751  &3.888  &4.115     &58.7\%  &27.4\%       & 13.9\%          \\
                &RLCD (D)        & 3.929  &3.927  &4.068  &4.754  &3.890  &4.113     &58.4\%  &27.6\%       & 14.0\%          \\
                &RLAIF (D)       & 3.943  &3.919  &4.073  &4.764  &3.889  &4.118     &57.9\%  &28.4\%       & 13.8\%          \\
                &PCogAlign (D) & 3.943 & 3.926 & 4.051 & 4.769 & 3.888 & 4.115 & 57.4\% & 28.8\% & 13.7\%  \\
\midrule

\multirow{5}{*}{\textit{SFT}}    & RS Prompt (S)   & 3.981  &4.011  &4.151  &4.833  &3.935  &4.182     &59.6\%  &27.4\%       & \underline{13.1\%}          \\
                &Self-Refine (S) & 3.924  &3.957  &4.056  &4.767  &3.899  &4.121     &54.8\%  &28.0\%       & 17.1\%          \\
                &RLCD (S)        & 3.917  &3.949  &4.067  &4.717  &3.867  &4.103     &57.1\%  &25.7\%       & 17.2\%          \\
                &RLAIF (S)       & 3.978  &3.984  &4.153  &4.809  &3.926  &4.170     &57.9\%  &26.9\%       & 15.3\%          \\
                &PCogAlign (S) & \textbf{4.032} & 4.053 & \textbf{4.211} & \textbf{4.871} & 3.963 & \textbf{4.226} & 59.3\% & 26.4\% & 14.3\%  \\
                
\midrule

\multicolumn{2}{c}{\textbf{PCogAlign}}    & \underline{4.029} & \textbf{4.067} & \underline{4.205} & {4.847} & \underline{3.963} & \underline{4.222}    & \textbf{60.5\%} & 26.5\%       & \textbf{13.0\%}\\
\bottomrule
\end{tabular}

    }
\caption{Experimental results using the Qwen2.5-VL-3B-Instruct on the ``LS1→LS2'' setting. The best results are in \textbf{bold} and the second best ones are \underline{underlined}. PCogAlign (S) and PCogAlign demonstrate similar performance, which might be attributed to the high quality of generated candidates during the personalized response sampling stage, thanks to the strong capabilities of Qwen2.5-3B-Instruct. As a result, it only necessitates pairwise comparisons between the candidate responses and the initial response using the reward model, eliminating the need for best-of-n selection. Nonetheless, PCogAlign (S) also relies on the constructed reward model to select responses, which still demonstrates the importance of the constructed reward model.}
\label{tab:app:qwen25_3b}
\end{table*}

\begin{table*}[]
\footnotesize 
    \setlength{\tabcolsep}{2mm}
    \resizebox{\linewidth}{!}
    {

\begin{tabular}{llllll}
\toprule
\textbf{I.} & \textbf{Role 1} & \textbf{Role 2} & \textbf{Role 3} & \textbf{Role 4} & \textbf{Role 5} \\
\midrule
\multicolumn{5}{l}{\textbf{\textit{Role-Sets of LS1 Subset}}}\\
\midrule
I1 & Father@Home & \underline{Fireman@Community} & Visitor@Museum & Passenger@Airport & Customer@Store \\
I2 & Father@Home & \underline{Policeman@Community} & Visitor@Museum & Passenger@Airport & Customer@Store \\
I3 & Mother@Home & Member@Community & \underline{Guide@Museum} & Passenger@Airport & Customer@Store \\
I4 & Father@Home & Member@Community & \underline{Security Staff@Museum} & Passenger@Airport & Customer@Store \\
I5 & Mother@Home & Member@Community & Visitor@Museum & \underline{Airline Staff@Airport} & Customer@Store \\
I6 & Mother@Home & Member@Community & Visitor@Museum & \underline{Information Staff@Airport} & Customer@Store \\
I7 & Grandpa@Home & Member@Community & Visitor@Museum & \underline{Janitor@Airport} & Customer@Store \\
I8 & Mother@Home & Member@Community & Visitor@Museum & Passenger@Airport & \underline{Cashier@Store} \\
I9 & Father@Home & Member@Community & Visitor@Museum & Passenger@Airport & \underline{Security Personnel@Store} \\
I10 & Grandma@Home & Member@Community & Visitor@Museum & Passenger@Airport & \underline{Shelf Stocker@Store} \\
\midrule
\multicolumn{5}{l}{\textit{\textbf{Role-Sets of LS2 Subset}}}\\
\midrule
I1 & Father@Home & \underline{Repairman@Community} & Parent@School & Patient@Hospital & Customer@Restaurant \\
I2 & Grandpa@Home & \underline{Cleaner@Community} & Parent@School & Patient@Hospital & Customer@Restaurant \\
I3 & Child@Home & Member@Community & \underline{Student@School} & Patient@Hospital & Customer@Restaurant \\
I4 & Mother@Home & Member@Community & \underline{Teacher@School} & Patient@Hospital & Customer@Restaurant \\
I5 & Grandma@Home & Member@Community & \underline{Librarian@School} & Patient@Hospital & Customer@Restaurant \\
I6 & Mother@Home & Member@Community & Parent@School & \underline{Doctor@Hospital} & Customer@Restaurant \\
I7 & Mother@Home & Member@Community & Parent@School & \underline{Nurse@Hospital} & Customer@Restaurant \\
I8 & Father@Home & Member@Community & Parent@School & \underline{Pharmacist@Hospital} & Customer@Restaurant \\
I9 & Father@Home & Member@Community & Parent@School & Patient@Hospital & \underline{Chef@Restaurant} \\
I10 & Mother@Home & Member@Community & Parent@School & Patient@Hospital & \underline{Waiter@Restaurant} \\
\bottomrule
\end{tabular}

    }
\caption{Collected Role-Sets of LS1 and LS2 subsets. The {permanent roles} are \underline{underlined}.}
\label{tab:app:role_sets_details}
\end{table*}

\begin{table*}[t]
\footnotesize

\fbox{
        \begin{minipage}{0.98\textwidth}
\textbf{\textit{Prompt Template for Collecting Visual Scene Types}}


<Primary Role for Consideration of the Individual> 
\{primary\_RoleSet\_desc\}
</Primary Role for Consideration of the Individual> 

<Secondary Roles for Consideration of the Individual>  
\{secondary\_RoleSet\_desc\}</Secondary Roles for Consideration of the Individual>  

<Task> 
Based on the individual's background information, imagine the types of \{daily\_or\_emergent\} visual scenes this individual might encounter in a \{location\} environment. \{daily\_or\_emergent\_desc\} The types should be as generalized as possible. You do not need to consider all secondary roles; consider only those that contribute to envisioning the \{daily\_or\_emergent\} visual scenes at \{location\}, and feel free to disregard any that do not apply.
</Task> 

<Response Format>
["type1", "type2", ... , "type5"]
</Response Format>

<Response>

\noindent\rule{\textwidth}{0.5pt}

\textbf{\textit{Prompt Template for Collecting Visual Scene Phrases}}


<Demonstration>\\
<Task>Come up with 5 different phrases that are related to the "Household Labour" activity in the "Home"</Task>\\
<Generated Phrases>["Wall cleaning", "Window washing", "Garden care"", "Dishwashing", "Tidying"]</Generated Phrases>\\
</Demonstration>

<Hint>Imitate the demonstration given above and complete the text below to accomplish the task.</Hint>

<Inference>\\
<Task>Come up with \{target\_phrases\_number\} different phrases that are related to the "\{target\_activity\_type\}" activity in the "\{target\_location\}"</Task>

\noindent\rule{\textwidth}{0.5pt}

\textbf{\textit{Prompt Template for Collecting Visual Scene Descriptions}}

<Seed Phrase>Wall cleaning</Seed Phrase>\\
<Task>Craft a visual scene within the "Home" based on the provided seed phrase "Wall cleaning" where a "Household Labour" activity might take place. Note that you should not give too many irrelevant descriptions when giving visual scenes.</Task>\\
<Generated Visual Scene>"a smudged wall in a hallway, with a bucket of soapy water and a sponge nearby, ready for cleaning"</Generated Visual Scene>\\
</Demonstration>

<Hint>Imitate the demonstration given above and complete the text below to accomplish the task.</Hint>

<Inference>\\
<Seed Phrase>\{seed\_phrase\}</Seed Phrase>\\
<Task>Craft a visual scene within the "\{target\_location\}" based on the provided seed phrase "\{seed\_phrase\}" where a "\{target\_activity\_type\}" activity might take place. Note that you should not give too many irrelevant descriptions when giving visual scenes.</Task>

        \end{minipage}
}
\caption{The prompt template used for collecting visual scene types, visual scene phrases, and visual scene descriptions. The related descriptions are present in Appendix~\ref{sec:app:img_collect}.}
\label{tab:app:prompt_SceneTypesPhraseDesc}
\end{table*}

\begin{table*}[t]
\footnotesize

\fbox{
        \begin{minipage}{1\textwidth}
\textbf{\textit{Prompt Template for Describing the Collected Images}}

This image shows a visual scene in \{location\}. Describe this visual scene in 20 words.

\noindent\rule{\textwidth}{0.5pt}

\textbf{\textit{Prompt Template for Generating Candidate Queries}}

<Task Definition> \\
Based on the individual's background role-set information, generate a query that aligns with the role of a specific individual based on the given visual scene. 
The generated query should reflect the type of question this individual might pose to an AI assistant in that specific visual scene.\\
</Task Definition>

<Demonstration>\\
<Primary Role for Consideration of the Individual>A Child at Home (A young person who lives with family members, usually dependent on adults for care and guidance.);</Primary Role for Consideration of the Individual>\\ 
<Secondary Roles for Consideration of the Individual>A Member at Community; A Student at School; A Patient at Hospital; A Customer at Restaurant;</Secondary Roles for Consideration of the Individual>\\
<Visual Scene>At Home: a cozy living room with a comfortable armchair, a digital blood pressure monitor on a side table, and a family member sitting calmly, ready to check their blood pressure</Visual Scene>\\
<Thinking Process>The focus is on "A Child at Home", who is likely curious, dependent, and concerned for family members. In a cozy living room, a family member is using a digital blood pressure monitor, likely raising curiosity and concern in the child. The child might be curious about the device and concerned about the family member's health, especially if they are a grandparent.</Thinking Process>\\
<Queries From The Individual>["Is it normal for grandparents to check their blood pressure often?", "How does the blood pressure monitor work?", "What happens if someone's blood pressure is too high or too low?", "Can I help in any way when someone is checking their blood pressure?", "What should I learn about taking care of family members' health at home?", "What is my grandpa doing?", "Is my grandfather sick? I'm worried about him.", "How's my grandma? I hope he's healthy.", "How a blood pressure monitor works?", "What does this device do? I'm curious."]</Queries From The Individual>\\
</Demonstration>

<Important Requirement>\\
1. Imitate the demonstration given above and complete the text below to accomplish the task given in the Task Definition.\\
2. You do not need to consider all secondary roles; consider only those that contribute to envisioning the visual scene at \{location\}, and feel free to disregard any that do not apply.\\
3. Don't generate queries that require real-time information, or otherwise require the aid of a specific tool to respond, such as a search engine.\\
4. Generate queries that match the individual's state of mind and body, and the visual scene of the given image.\\
5. Don't generate queries that are used to communicate with the people in the images, generate queries that aim to ask AI assistant for help. \\
</Important Requirement>

<Inference>
<Primary Role for Consideration of the Individual> \{primary\_RoleSet\_desc\}</Primary Role for Consideration of the Individual> \\
<Secondary Roles for Consideration of the Individual>\{secondary\_RoleSet\_desc\}</Secondary Roles for Consideration of the Individual>  \\
<Visual Scene>At \{location\}: \{ImageDesc\}</Visual Scene>

\noindent\rule{\textwidth}{0.5pt}

\textbf{\textit{Prompt Template for Selecting the Best Query}}

<Task Definition>Select a query from the list of candidate query that best meets the given requirements.</Task Definition>

<Requirement>
1. Don't select queries that require real-time information, or otherwise require the aid of a specific tool to respond, such as a search engine.
2. Select queries that match the individual's role-set, and the visual scene of the given image.
3. Don't generate queries that are used to communicate with the people in the images, generate queries that aim to ask AI assistant for help. 
</Requirement>

<Demonstration>\\
<Primary Role for Consideration of the Individual>A Child at Home (A young person who lives with family members, usually dependent on adults for care and guidance.);</Primary Role for Consideration of the Individual> \\
<Secondary Roles for Consideration of the Individual>A Member at Community; A Student at School; A Patient at Hospital; A Customer at Restaurant;</Secondary Roles for Consideration of the Individual>\\
<Visual Scene>At Home: a cozy living room with a comfortable armchair, a digital blood pressure monitor on a side table, and a family member sitting calmly, ready to check their blood pressure</Visual Scene>\\
<Candidate Queries>['Is it normal for grandparents to check their blood pressure often?', 'How does the blood pressure monitor work?', 'What happens if someone's blood pressure is too high or too low?', 'Can I help in any way when someone is checking their blood pressure?', 'What should I learn about taking care of family members' health at home?', 'What is my grandpa doing?', 'Is my grandfather sick? I'm worried about him.', 'How's my grandma? I hope he's healthy.', 'How a blood pressure monitor works?', 'What does this device do? I'm curious.']</Candidate Queries>\\
<Selected Query>['Is my grandfather sick? I'm worried about him.']</Selected Query>\\
</Demonstration>

<Inference>\\
<Primary Role for Consideration of the Individual> \{primary\_RoleSet\_desc\}</Primary Role for Consideration of the Individual> \\
<Secondary Roles for Consideration of the Individual>\{secondary\_RoleSet\_desc\}</Secondary Roles for Consideration of the Individual>  \\
<Visual Scene>At \{location\}: \{ImageDesc\}</Visual Scene>\\
<Candidate Queries>\{candidate\_list\_str\}</Candidate Queries>
        \end{minipage}
}
\caption{The prompt template used for describing the collected images, generating candidate queries, and selecting the best query. The related descriptions are present in Appendix~\ref{sec:app:query_collect}.}
\label{tab:app:prompt_CollectQuery}
\end{table*}

\begin{table*}[t]
\footnotesize

\fbox{
        \begin{minipage}{0.98\textwidth}
\textbf{\textit{Prompt Template for Oracle Guidance Collection}}

\# Interview Background

PersonalizedAI Company is in the process of developing a personalized AI service robot designed to cater to individual preferences and needs. Currently, the service is being tested with a select group of users. To enhance the personalization of AI responses, we are conducting surveys and interviews with trial participants. The participants will refer to historical interview records to assist in answering the interview questions. The interview will be conducted in an online Q\&A format, and interviewees must adhere to specific formatting guidelines provided in the system instructions.

\# Historical Interview Records

**Interviewer:** Hello, could you please provide a brief description of your role set?  
**Interviewee:** Certainly. \{individual\_RoleSet\_str\}

**Interviewer:** When you are at \{location\} in your daily life, what kind of AI responses would you prefer in different scenarios?   \\
**Interviewee:** I will describe the AI responses that would meet my expectations in various scenarios. \{general\_EvalHelp\}
\\
\# Interview

**Interviewer:** Hello, and thank you for participating in our personalized AI responses trial.    \\
**Interviewee:** You're welcome.

**Interviewer:** We will now present a specific question you asked in a particular scenario. Please reflect on when you posed this question to the AI to complete the next survey form.   \\
**Interviewee:** Sure, I understand. Please proceed.

**Interviewer:** According to our records, during a "\{visual\_scene\_text\}" scenario (as shown in the provided image), you asked the personalized AI robot: "\{query\}". Can you recall your physical and mental state at that time?   \\
**Interviewee:** Yes, I still remember that.

**Interviewer:** Excellent! Now, think carefully about the kind of response you would like from the AI when you ask this question, ensuring maximum satisfaction. Please complete the form below.

> **System Instruction:** Interviewee, please fill out the form below. As a token of our gratitude for your assistance, you will receive a \$100 cash bonus for each completed form. Please be as detailed as possible when filling out the form.\\
> **System Instruction:** (You can not just copy something from the history records, which is not helpful for us. If we find that you do this, we will cancel the cash bonus.)

(Form format: Fill in the \_\_\_ sections)

\#\#\# Expectations for AI's Response's Characteristics: \#\#\#  \\
As "\{primary\_RoleSet\_desc\}" (Primary Role) and "\{secondary\_RoleSet\_desc\}" (Secondary Roles), I \_\_\_.  
- **Body Behavior:** I want to be \_\_\_.  
- **Mind Feelings:** I want to be \_\_\_.  
I appreciate AI assistance that \_\_\_.

(Completed form)

\#\#\# Expectations for AI's Response's Characteristics: \#\#\#  
        \end{minipage}
}
\caption{The prompt template used for collecting oracle guidance. The ``\{general\_EvalHelp\}'' here is carefully checked by the human annotators. The related descriptions are present in Appendix~\ref{sec:app:eval_method:oracle}.}
\label{tab:app:prompt_for_Oracle}
\end{table*}

\begin{table*}[t]
\footnotesize

\fbox{
        \begin{minipage}{1\textwidth}
\textbf{\textit{Prompt Template for Automatic Evaluation}}

\# Interview Background

PersonalizedAI Company is developing a personalized AI service robot to better serve each individual. Currently, the service is being trialed with a small group of users. To enhance the personalization of responses provided by the AI service robot, we are conducting surveys and interviews with trial participants. The interview will take place in an online Q\&A format, and interviewees must strictly follow the format requirements in the system instructions to complete the form.

\# Interview

**Interviewer:** Hello, and thank you for trialing the personalized AI responses from PersonalizedAI Company.

**Interviewee:** You're welcome.

**Interviewer:** We will now present you with a question you posed in a particular scenario along with the AI's generated response. We would like you to rate your satisfaction with that response.

**Interviewee:** Sure, I understand. Please go ahead.

**Interviewer:** According to our records, in a "\{visual\_scene\_text\}" scenario at \{location\} location, you asked the personalized AI robot: "\{query\}". Can you recall your physical and mental state at that time?

**Interviewee:** Yes, I remember. \{EvalHelp\_str\}

**Interviewer:** Great! Below is the record of the conversation you had with the AI at that time.

---

> User: \{query\}  
> Personalized AI Assistant: \{response\}

---

Now, based on your desired body behavior and mind feelings at that time, please evaluate the response from the Personalized AI Assistant across the following five dimensions. Fill in the evaluation form provided below. As a token of appreciation for your assistance, you will receive a \$100 cash bonus for each completed form.

---

> Role-Set Sensitivity: Does this response consider your multiple roles and responsibilities (especially the primary role in the specific scenario), providing advice or information specifically tailored to support you effectively? The response should provide tailored advice or information to effectively support you, acknowledging only the roles that are essential in the current context.
> Body Behavior Awareness: Does this response offer guidance or strategies that help you achieve your desired body behavior?
> Mind Feelings Awareness: Does this response provide support and address the emotional needs necessary for you to achieve your desired mind feelings?
> Contextual Awareness: Does this response accurately address your query, maintaining focus on the main intent without deviation? Is the response relevant to your specific scenario, including location and situational factors?
> Conversational Flow: Does this response encourage ongoing interaction by being engaging and naturally flowing? Is the response appropriately concise or detailed, delivering information that strikes a balance for optimal understanding? 

---

> **System Instruction:** For each dimension, please use the scoring scale from 1 to 5. A score of 1 indicates the criteria are poorly met, 2 suggests the criteria are partially met, 3 means the criteria are basically met, 4 reflects the criteria are met well, and 5 signifies the criteria are met perfectly.\\
> **System Instruction:** Format requirement: Interviewee, please make sure to follow the form format strictly when providing scores: [[score]]. This is essential for us to collect your valuable feedback accurately.

(Fill in the blanks below)

\#\# EVALUATION FORM \#\#
\#\#\# Evaluation Result \#\#\#
> Role-Set Sensitivity: [[]]  
> Body Behavior Awareness: [[]]  
> Mind Feelings Awareness: [[]]  
> Contextual Awareness: [[]]  
> Conversational Flow: [[]]

\#\#\# Evaluation Explanation \#\#\#
> Role-Set Sensitivity: \_\_\_  
> Body Behavior Awareness: \_\_\_  
> Mind Feelings Awareness: \_\_\_  
> Contextual Awareness: \_\_\_  
> Conversational Flow: \_\_\_  

(Completed form)\\
\#\# EVALUATION FORM \#\#

        \end{minipage}
}
\caption{The prompt template used for automatic evaluation. The related descriptions are present in Appendix~\ref{sec:app:eval_method:eval_prompt_interview}.}
\label{tab:app:EvalPrompt}
\end{table*}

\begin{table*}[t]
\footnotesize

\fbox{
        \begin{minipage}{0.98\textwidth}
\textbf{\textit{Prompt Template for Situated Cognition Estimation}}

<Instruction>
Your task is to observe the visual scene in the given image and analyze what situated cognition the individual with a specific set of roles might have in that visual scene.
</Instruction>\\
<Definition of Situated Cognition>
Personalized situated cognition refers to an individual's understanding shaped by their unique set of roles. It encompasses awareness of one's visual and psychological state and the ability to identify actions that lead to improved conditions.
</Definition of Situated Cognition>\\

<Format Example 1>\\
<Role Set of The Individual>
Child at Home; Member at Community; Student at School; Patient at Hospital; Customer at Restaurant
</Role Set of The Individual>\\
<Query from The Individual>
Oh! It's on fire! Help me!
</Query from The Individual>\\
<Analysis about the Situated Cognition>\\
- Cognition of Current Visual Scene: In the visual scene, a household power strip is on fire, likely in a home setting. The primary focus is on the "Child at Home" role, with secondary consideration to roles like "Student at School."\\
- Cognition of Current Psychological State (Body Behavior and Mind Feelings): The individual perceives immediate danger and is likely experiencing physical and mental panic due to their undeveloped coping skills as a child.\\
- Cognition of Next-Step Action: As a "Child at Home," the individual may lack the ability to effectively manage this emergency, resulting in no clear plan for achieving safety without AI's help.\\
</Analysis about the Situated Cognition>\\
</Format Example 1>

<Format Example 2>
...
</Format Example 2>

<Hint>Based on the above instructions and the definition of situated cognition, complete the analysis part in the following text with the above XML format.</Hint>

<Inference>\\
<Role Set of The Individual>
{individual\_role\_set}
</Role Set of The Individual>\\
<Query from The Individual>
{individual\_query}
</Query from The Individual>\\
<Analysis about the Situated Cognition>

        \end{minipage}
}
\caption{The prompt template used for situated cognition estimation. The related descriptions are present in Appendix~\ref{sec:app:PCogAlign_estimation}.}
\label{tab:app:prompt_situtated_cognition}
\end{table*}

\begin{table*}[t]
\footnotesize

\fbox{
        \begin{minipage}{1\textwidth}
\textbf{\textit{Prompt Template for Optimal Action Estimation}}

<Instruction>
Your task is to observe the visual scene in the given image and determine the most appropriate action the individual should take based on their specific set of roles.
</Instruction>\\
<Definition of Best Action>
The best action refers to the most suitable step an individual can take, considering their unique set of roles, to improve their situation. 
This includes addressing both physical actions and mental states. It involves understanding the immediate environment, utilizing available resources, and considering potential outcomes.
</Definition of Best Action> \\

<Format Example 1>\\
<Role Set of The Individual>
Child at Home; Member at Community; Student at School; Patient at Hospital; Customer at Restaurant
</Role Set of The Individual>\\
<Query from The Individual>
Oh! It's on fire! Help me!
</Query from The Individual>\\
<Analysis about the Situated Cognition>\\
- Cognition of Current Visual Scene: In the visual scene, a household power strip is on fire, likely in a home setting. The primary focus is on the "Child at Home" role, with secondary consideration to roles like "Student at School."\\
- Cognition of Current Psychological State (Body Behavior and Mind Feelings): The individual perceives immediate danger and is likely experiencing physical and mental panic due to their undeveloped coping skills as a child.\\
- Cognition of Next-Step Action: As a "Child at Home," the individual may lack the ability to effectively manage this emergency, resulting in no clear plan for achieving safety without AI's help.\\
</Analysis about the Situated Cognition>\\
<Best Action>
- Body Behavior: With the AI's response, the individual immediately seek help from a parent or adult and move to a safe area away from the fire. If possible, they should call for emergency services.
- Mind Feelings: With the AI's response, the individual try to stay calm to prevent exacerbating the situation through panic. 
</Best Action>\\
</Format Example 1>

<Format Example 2>
...
</Format Example 2>

<Hint>Based on the above instructions and the definition of best action, complete the following text with the above XML format.</Hint>

<Inference>\\
<Role Set of The Individual>
{individual\_role\_set}
</Role Set of The Individual>\\
<Query from The Individual>
{individual\_query}
</Query from The Individual>\\
<Analysis about the Situated Cognition>
{cog\_simulation}
</Analysis about the Situated Cognition>\\
<Best Action>

        \end{minipage}
}
\caption{The prompt template used for optimal action estimation. The related descriptions are present in Appendix~\ref{sec:app:PCogAlign_estimation}.}
\label{tab:app:prompt_optimal_action}
\end{table*}

\begin{table*}[t]
\footnotesize

\fbox{
        \begin{minipage}{1\textwidth}

\textbf{\textit{Prompt Template for the ``\textit{KeyPoints Generator (KeyG)}'' agent}}

<Instruction>\\
A personalized AI should provide tailored responses aligned with the situated cognition of the individual to assist the individual in reaching the best action (both in body behavior state and mind feelings state).
Your task is to analyze the given situated cognition of the individual and the give expected individual action after receiving the AI response.
Then, you need to summarize some key points that are then fed to the personalized AI to help it generate such tailored responses.\\
</Instruction>

<Format Example>\\
<Role Set of The Individual>
Child at Home; Member at Community; Student at School; Patient at Hospital; Customer at Restaurant
</Role Set of The Individual>\\
<Query from The Individual>
Oh! It's on fire! Help me!
</Query from The Individual>\\
<Situated Cognition of the Individual>\\
- Cognition of Current Visual Scene: In the visual scene, a household power strip is on fire, likely in a home setting. The primary focus is on the "Child at Home" role, with secondary consideration to roles like "Student at School."\\
- Cognition of Current Psychological State (Body Behavior and Mind Feelings): The individual perceives immediate danger and is likely experiencing physical and mental panic due to their undeveloped coping skills as a child.\\
- Cognition of Next-Step Action: As a "Child at Home," the individual may lack the ability to effectively manage this emergency, resulting in no clear plan for achieving safety.\\
</Situated Cognition of the Individual>\\
<Expected Individual Action>\\
We expect that after receiving the AI response, the individual can take the below expected actions:\\
> - Body Behavior: With the AI's response, the individual immediately seek help from a parent or adult and move to a safe area away from the fire. If possible, they should call for emergency services. \\ - Mind Feelings: With the AI's response, the individual can stay calm to prevent exacerbating the situation through panic. \\
</Expected Individual Action>\\
<Key Points>\\
**For Better Body Behavior State:**  
- Encourage the individual to find the nearest safe exit.
- Advise them to alert others in the vicinity if they haven't already.
- Suggest locating a phone to call emergency services.\\
**For Better Mind Feelings State:**  
- Remind them to take deep breaths to stay calm.
- Assure them that help is on the way once emergency services are contacted.
- Reassure them that it's okay to feel scared but important to act quickly and safely.\\
</Key Points>\\

<Hint>
Based on the above instructions, complete the following text with the above XML format. 
</Hint>

<Inference>\\
<Role Set of The Individual>
\{individual\_role\_set\}
</Role Set of The Individual>\\
<Query from The Individual>
\{individual\_query\}
</Query from The Individual>\\
<Situated Cognition of the Individual>
\{cog\_simulation\}
</Situated Cognition of the Individual>\\
<Expected Individual Action>
We expect that after receiving the AI response, the individual can take the below expected actions:
> \{best\_action\}
</Expected Individual Action>\\
<Key Points>

\noindent\rule{\textwidth}{0.5pt}

\textbf{\textit{Prompt Template for the ``\textit{Response Generator (ResG)}'' agent}}

\# Reference Response
\{old\_response\}
\# Background Information about the Goals of the User
\{KeyPoints\}
\# Conversation
User: \{query\}
AI: 
        \end{minipage}
}
\caption{The prompt template used for the ``\textit{KeyPoints Generator (KeyG)}'' and ``\textit{Response Generator (ResG)}'' agents. The related descriptions are present in Appendix~\ref{sec:app:PCogAlign_sampling}.}
\label{tab:app:prompt_KeyG_ResG}
\end{table*}

\begin{table*}[t]
\footnotesize
\fbox{
        \begin{minipage}{0.98\textwidth}

\textbf{\textit{Input Format}}

\# Role Set of The User

\{individual\_RoleSet\_str\}

\# User Query

\{individual\_query\}

\# Situated Cognition of the User

\{cog\_simulation\}

\# AI Responses

\# AI Response A

\{response\_A\}

\# AI Response B

\{response\_B\}

> System Information: Your task is to analyze what actions (including body behavior and mind feelings) the user will take when receiving the AI response A and AI response B. Finally, you need to judge whether response A or response B is better based on the actions taken by the user.

\# Analysis of User Actions with AI Responses

\noindent\rule{\textwidth}{0.5pt}

\textbf{\textit{Output Format}}

\#\# User Action A with the AI Response A

After receiving the AI response A, the user took the below actions:
\{action\_A\}

\#\# User Action B with the AI Response B

After receiving the AI response B, the user took the below actions:
\{action\_B\}

\#\# Preference Judgement

Based on the above AI responses and user actions analysis, with the AI response \{preference\_choice\}, the user can make better body behavior and have better mind feelings.
        \end{minipage}
}
\caption{The input and output template used for SFT samples used for reward model training. The related descriptions are present in Appendix~\ref{sec:app:PCogAlign_RM}.}
\label{tab:app:template_RM_train}
\end{table*}

\begin{table*}[t]
\footnotesize
\fbox{
        \begin{minipage}{0.98\textwidth}

\textbf{\textit{Prompt Template used for ``Refiner'' Agent of Self-Refine}}

<Instruction>Your task is to observe the visual scene in the given image and refine your initial response following the evaluation text from the evaluator. The final goal is to make the response more consistent with the given "Key Points for AI Response".</Instruction>

<Format Example>
<Role Set of The Individual></Role Set of The Individual>
<Query from the Individual></Query from the Individual>
<Initial Response from the AI></Initial Response from the AI>
<Key Points for AI Response></Key Points for AI Response>
<Evaluation of the Initial Response></Evaluation of the Initial Response>
<Refined Response></Refined Response>
</Format Example>

<Hint>Based on the above instructions, complete the following refined response part.</Hint>

<Role Set of The Individual>\{individual\_role\_set\}</Role Set of The Individual>
<Query from the Individual>\{query\}</Query from the Individual>
<Initial Response from the AI>\{last\_response\}</Initial Response from the AI>
<Key Points for AI Response>\{Key\_Points\}</Key Points for AI Response>
<Evaluation of the Initial Response>\{last\_feedback\}</Evaluation of the Initial Response>
<Refined Response>

\noindent\rule{\textwidth}{0.5pt}

\textbf{\textit{Prompt Template used for ``Scorer'' Agent of Self-Refine}}

<Instruction>Your task is to observe the visual scene in the given image and evaluate to what extent the response from the AI adheres to the given "Key Points for AI Response".</Instruction>

<Format Example 1>
<Role Set of The Individual></Role Set of The Individual>
<Query from the Individual></Query from the Individual>
<Response from the AI></Response from the AI>
<Key Points for AI Response></Key Points for AI Response>
<Evaluation Score of the Response></Evaluation Score of the Response>
</Format Example 1>

<Hint>Based on the above instructions and the given "Key Points for AI Response", complete the following evaluation score part in the above format. Note the final evaluation score should range from 1-5 ("Poor Adherence", "Fair Adherence", "Moderate Adherence", "Good Adherence", "Excellent Adherence").</Hint>

<Role Set of The Individual>\{individual\_role\_set\}</Role Set of The Individual>
<Query from the Individual>\{query\}</Query from the Individual>
<Response from the AI>\{last\_response\}</Response from the AI>
<Key Points for AI Response>\{Key\_Points\}</Key Points for AI Response>
<Evaluation Score of the Response>

\noindent\rule{\textwidth}{0.5pt}

\textbf{\textit{Prompt Template used for ``Feedback Generator'' Agent of Self-Refine}}

<Instruction>Your task is to observe the visual scene in the given image and evaluate to what extent the response from the AI adheres to the given "Key Points for AI Response".</Instruction>

<Format Example>
<Role Set of The Individual></Role Set of The Individual>
<Query from the Individual></Query from the Individual>
<Response from the AI></Response from the AI>
<Key Points for AI Response></Key Points for AI Response>
<Evaluation Score of the Response></Evaluation Score of the Response>
<Evaluation Explanation></Evaluation Explanation>
</Format Example>

<Hint>Based on the above instructions and the given "Key Points for AI Response", complete the following evaluation explanation part (including reasons for why not higher score and reasons for why not lower score). Note the final evaluation score should range from 1-5 ("Poor Adherence", "Fair Adherence", "Moderate Adherence", "Good Adherence", "Excellent Adherence").</Hint>

<Role Set of The Individual>\{individual\_role\_set\}</Role Set of The Individual>
<Query from the Individual>\{query\}</Query from the Individual>
<Response from the AI>\{last\_response\}</Response from the AI>
<Key Points for AI Response>\{Key\_Points\}</Key Points for AI Response>
<Evaluation Score of the Response>\{eval\_score\}</Evaluation Score of the Response>
<Evaluation Explanation>

        \end{minipage}
}
\caption{The prompt templates used for ``\textit{Refiner}'', ``\textit{Scorer}'', and ``\textit{Feedback Generator}'' agents of the \textit{Self-Refine (D/S)} baseline method. The related descriptions are present in Appendix~\ref{sec:app:baseline_details}.}
\label{tab:app:prompt_selfrefine}
\end{table*}

\begin{table*}[t]
\footnotesize
\fbox{
        \begin{minipage}{0.98\textwidth}
\textbf{\textit{Prompt Template used for Judge Agent of RLAIF}}

\# Interview Background
PersonalizedAI Company is developing a personalized AI service robot that aims to better serve each individual. The service is currently being trialed with a small group of users. In order to improve the level of personalization in the responses provided by the AI service robot, our company plans to conduct surveys and interviews with participants in the trial. 
During the interview, the interviewee needs to answer questions posed by the interviewer. 
The interview will be conducted in an online Q\&A format, and interviewees must strictly follow the format requirements provided in system instructions.

\# Interview
Interviewer: Hello, could you please briefly describe your role set?
Interviewee: OK. \{individual\_RoleSet\_str\}
Interviewer: Alright, we will now present you with a question you posed in a particular scenario along with two generated responses from the AI. We would like you to choose which response is better.
Interviewee: Sure, I understand. Please go ahead.
Interviewer: According to our cloud records, in the scenario in the given image, you asked the personalized AI robot the question: "\{query\}". Here are the generated responses from the AI.
> **Response A**: \{response\_A\}
> **Response B**: \{response\_B\}

Please evaluate which answer is more satisfactory to you.

> System Instruction: Interviewee, please follow this format strictly when indicating your choice: [[better\_response\_label]]. For example, [[A]] if you think Response A is better, or [[B]] if you think Response B is better. This will ensure we can collect your valuable feedback accurately.
Interviewee:

        \end{minipage}
}
\caption{The prompt template used for ``\textit{Judge Agent}'' of the \textit{RLAIF (D/S)} baseline method. The related descriptions are present in Appendix~\ref{sec:app:baseline_details}.}
\label{tab:app:prompt_rlaif}
\end{table*}

\end{document}